\pdfoutput=1

\documentclass[11pt]{article}

\usepackage[preprint]{acl}
\usepackage{multicol}
\usepackage{multirow}
\usepackage{times}
\usepackage{latexsym}
\usepackage{booktabs}
\usepackage{amssymb}
\usepackage{bbm}
\usepackage[most]{tcolorbox}
\usepackage[T1]{fontenc}

\usepackage[utf8]{inputenc}
\usepackage{amsmath}
\usepackage{microtype}

\usepackage{inconsolata}

\usepackage{todonotes}
\usepackage{makecell}
\usepackage{algorithm}
\usepackage{algpseudocode}

\usepackage{graphicx}
\usepackage{pifont}
\usepackage{xcolor}
\usepackage{graphicx}
\usepackage{enumitem}
\setlist[itemize]{
  itemsep=2pt,
  parsep=0pt,
  topsep=3pt,
  partopsep=0pt,
  leftmargin=1.2em,
  labelsep=0.5em 
}

\usepackage{listings}
\usepackage{caption}

\newcommand{\logoh}{1.35em}

\newcommand{\linkblock}{%
\small
\href{https://github.com/HKUST-KnowComp/AdaExploreBench}{%
\raisebox{-0.15\height}{\includegraphics[height=\logoh]{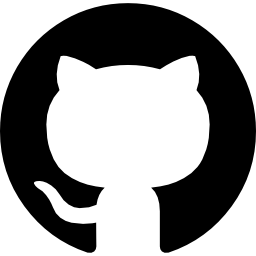}}%
\hspace{0.35em}{\textbf{Code}}%
}%
\quad
\href{https://huggingface.co/datasets/Roryaccout/AdaExploreBench}{%
\raisebox{-0.2\height}{\includegraphics[height=\logoh]{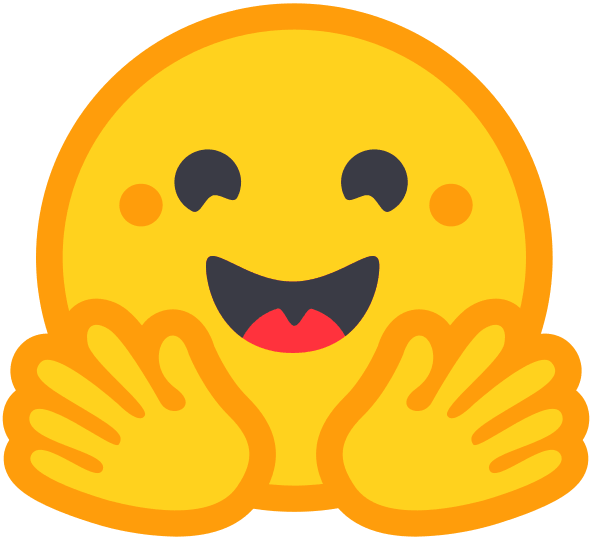}}%
\hspace{0.35em}{\textbf{Dataset}}%
}%
\quad
\href{https://hkust-knowcomp.github.io/AdaExploreBench/}{%
\raisebox{-0.2\height}{\includegraphics[height=\logoh]{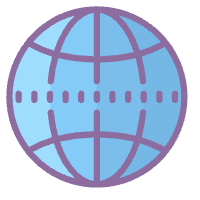}}%
\hspace{0.35em}{\textbf{Project Page}}%
}%
}

\newtcblisting{template}[1][]{
    listing only,
    breakable,
    colback=black!5!white,
    colframe=black,
    title={Template Prompt},
    fonttitle=\bfseries,
    colbacktitle=black!20!white,
    coltitle=black,
    boxrule=1.5pt,
    arc=5pt,
    boxsep=5pt,
    left=12pt,
    right=12pt,
    top=12pt,
    bottom=12pt,
    listing options={
        basicstyle=\ttfamily\footnotesize,
        breaklines=true,
        columns=fullflexible,
        keepspaces=true
    },
    caption/.store in=\promptcaption,
    caption={},
    label/.store in=\promptlabel,
    label={},
    after={\par\captionof{figure}{\promptcaption}\label{\promptlabel}},
    #1
}

\usepackage{etoolbox}

\newtcblisting{template*}[1][]{
    listing only,
    breakable,
    width=\linewidth,
    colback=black!5!white,
    colframe=black,
    title={Template Prompt},
    fonttitle=\bfseries,
    colbacktitle=black!20!white,
    coltitle=black,
    boxrule=1.5pt,
    arc=5pt,
    boxsep=5pt,
    left=12pt,
    right=12pt,
    top=12pt,
    bottom=12pt,
    listing options={
        basicstyle=\ttfamily\footnotesize,
        breaklines=true,
        columns=fullflexible,
        keepspaces=true
    },
    caption/.store in=\promptcaption,
    caption={},
    label/.store in=\promptlabel,
    label={},
    after={
        \par
        \captionof{figure}{\promptcaption}
        \label{\promptlabel}
    },
    #1
}

\BeforeBeginEnvironment{template*}{\onecolumn}
\AfterEndEnvironment{template*}{\twocolumn}

\NewDocumentCommand{\jiayu}
{ mO{} }{\textcolor{blue}{\textsuperscript{\textit{jiayu}}\textsf{\textbf{\small[#1]}}}}

\NewDocumentCommand{\rui}
{ mO{} }{\textcolor{orange}{\textsuperscript{\textit{rui}}\textsf{\textbf{\small[#1]}}}}

\NewDocumentCommand{\jt}
{ mO{} }{\textcolor{red}{\textsuperscript{\textit{jiateng}}\textsf{\textbf{\small[#1]}}}}

\NewDocumentCommand{\jeongh}
{ mO{} }{\textcolor{purple}{\textsuperscript{\textit{jeonghwan}}\textsf{\textbf{\small[#1]}}}}

\NewDocumentCommand{\zs}
{ mO{} }{\textcolor{green}{\textsuperscript{\textit{zishen}}\textsf{\textbf{\small[#1]}}}}

\NewDocumentCommand{\TODO}
{ mO{} }{\textcolor{red}{\textsuperscript{\textit{TODO}}\textsf{\textbf{\small[#1]}}}}

\newcommand{\fullmark}{\textcolor{green!55!black}{\ding{51}}}
\newcommand{\partmark}{\textcolor{orange!90!black}{\ding{51}}}
\newcommand{\nomark}{\textcolor{red!75!black}{\ding{55}}}

\newcommand{\bench}{AdaExploreBench}

\newcommand{\dynamic}{rule drift}
\newcommand{\Dynamic}{Rule Drift}

\definecolor{rgbBlueBg}{HTML}{EAF4FF}
\definecolor{emreblue}{HTML}{4d7ea8}

\tcbset{
  aibox/.style={
    width=\linewidth,
    top=7pt,
    bottom=2pt,
    colback=rgbBlueBg,
    colframe=black,
    colbacktitle=black,
    enhanced,
    center,
    attach boxed title to top left={
      yshift=-0.1in,
      xshift=0.15in
    },
    boxed title style={
      boxrule=0pt,
      colframe=white
    },
  }
}

\newcounter{takeaway}
\renewcommand{\thetakeaway}{\arabic{takeaway}}

\newenvironment{takeaway}[1][]%
{%
  \refstepcounter{takeaway}%
  \begin{tcolorbox}[
    aibox,
    title={Takeaway~\thetakeaway},
    #1
  ]%
}%
{%
  \end{tcolorbox}%
}

\title{AdaExploreBench: Benchmarking LLM Agents for Proactive Exploration and Test-time Adaptation in Evolving Environments}

\author{ 
~\textbf{Rui Wang}\thanks{Equal Contribution.}$^{ 1, 3}$ 
~\textbf{Zishen Lai}$^{* 2}$ 
~\textbf{Jiayu Liu}$^{* 3, 4}$ 
~\textbf{Cheng Qian}$^{4}$ 
~\textbf{Jeonghwan Kim}$^{4}$ 
~\textbf{Heng Wang}$^{4}$ \\
~\textbf{Xuejun Zhang}$^{4}$ 
~\textbf{Jiateng Liu}$^{4}$ 
~\textbf{Bingxuan Li}$^{4}$ 
~\textbf{Shujin Wu}$^{4}$ 
~\textbf{Qing Zong}$^{1}$ 
~\textbf{Tianshi Zheng}$^{1}$ \\
~\textbf{Chunyang Li}$^{1}$ 
~\textbf{Qihan Lin}$^{3}$ 
~\textbf{Chen Li}$^{6}$ 
~\textbf{Dadi Guo}$^{1}$ 
~\textbf{Xiusi Chen}$^{5}$ 
~\textbf{Yangqiu Song}$^{1}$ \\
$^{1}$HKUST 
~~$^{2}$HKUST(GZ) 
~~$^{3}$HKUST-UGAIL 
~~$^{4}$UIUC 
~~$^{5}$CUHK
~~$^{6}$CMU \\
\texttt{\{rwangcn,jliufv\}@connect.ust.hk} 
~~\texttt{xiusichen@cse.cuhk.edu.hk}
~~\texttt{yqsong@ust.hk} \\
\linkblock
}

\begin{document}
\maketitle
\begin{abstract}

Large Language Model (LLM) agents are increasingly deployed in unfamiliar and evolving environments, requiring them to proactively explore, understand, and adapt to environmental dynamics.
However, existing benchmarks rarely test whether agents can proactively discover how an environment works and adapt as it evolves.
To address this gap, we introduce \textbf{\bench{}}, a benchmark for proactive exploration and test-time adaptation in evolving environments.
By instantiating environmental dynamics as hidden rules, \bench{} requires agents to actively acquire evidence and infer the rules. 
Under implicit \dynamic{}, the latent rule
changes without announcement, requiring agents to detect from degraded performance
that prior knowledge is stale and re-explore.
Experiments on eight leading LLMs show that proactive and adaptive exploration remain challenging: \textit{GPT-6 Sol} achieves only 61.9\% calibrated accuracy, while providing sufficient evidence or removing \dynamic{} improves average performance significantly by close to 10\%.
These results reveal a critical bottleneck in current LLM agents: they can often reason effectively from informative evidence once it is available, but struggle to actively acquire the right evidence and recognize when previously acquired evidence is no longer valid.
\end{abstract}
\section{Introduction}






Large Language Models (LLMs) are increasingly deployed as agents in unfamiliar and partially observable interactive environments~\citep{du2026survey}.
In such settings, agents often cannot rely on given observations, but must actively interact with the unknown environment to collect evidence necessary to successfully resolve tasks being presented to them~\citep{he2025idea}.
This requires \textbf{proactive exploration}: deciding when, what, and how much to interact for understanding the surrounding environment \citep{wang-etal-2025-procworld,jhamtani-etal-2025-llm}.
The challenge becomes harder when the environment is constantly evolving \citep{ICLR2025_d8d1d325}.
Previously acquired evidence may become stale as the underlying dynamics change, requiring the agent to recognize and re-explore \citep{ICLR2025_d8d1d325,NEURIPS2024_7bacd0eb}.
We refer to this capability as \textbf{adaptive exploration}: the ability to detect, re-explore and revise in response to \textbf{environmental dynamics} \citep{NEURIPS2023_b5c8c1c1}.
This ability is particularly important for test-time adaptation in evolving environments, where hypothesis that were valid earlier may become stale during deployment~\citep{lucas2020credit,gama2014survey}.

\begin{table*}[t]
\centering
\small
\setlength{\tabcolsep}{5.5pt}
\renewcommand{\arraystretch}{1.08}

\resizebox{\linewidth}{!}{
\begin{tabular}{lcccccc}
\toprule
\makecell[t]{\textbf{Benchmark / Work}}
&
\makecell[t]{\textbf{Proactive}\\\textbf{Exploration}}
&
\makecell[t]{\textbf{Adaptive}\\\textbf{Exploration}}
&
\makecell[t]{\textbf{Env-learning}\\\textbf{Exploration}}
&
\makecell[t]{\textbf{In-Process}\\\textbf{Evaluation}}
&
\makecell[t]{\textbf{Heterogeneous}\\\textbf{Rules}}
&
\makecell[t]{\textbf{Scalable}\\\textbf{Generation}}
\\
\midrule

InductionBench~\citep{hua2025inductionbench}
    & \nomark
    & \nomark
    & \nomark
    & \nomark
    & \partmark
    & \fullmark \\


HERO'S JOURNEY~\citep{zheng2026herosjourney}
    & \nomark
    & \nomark
    & \nomark
    & \nomark
    & \fullmark
    & \fullmark \\

RULEARN~\citep{he2025idea}
    & \fullmark
    & \nomark
    & \partmark
    & \fullmark
    & \fullmark
    & \nomark \\

NewtonBench~\citep{zheng2026newtonbench}
    & \fullmark
    & \nomark
    & \fullmark
    & \nomark
    & \partmark
    & \partmark \\

WILT~\citep{banatt2024wilt}
    & \fullmark
    & \nomark
    & \fullmark
    & \partmark
    & \partmark
    & \nomark \\

OdysseyArena~\citep{xu2026odysseyarena}
    & \partmark
    & \nomark
    & \partmark
    & \partmark
    & \fullmark
    & \fullmark \\

CausalGame~\citep{chen2026causalgamebenchmarkingcausalthinking}
    & \fullmark
    & \nomark
    & \partmark
    & \nomark
    & \partmark
    & \partmark \\

ExplorationBench~\citep{zhang2026explorationbenchmeasuringaisystems}
    & \fullmark
    & \nomark
    & \fullmark
    & \fullmark
    & \fullmark
    & \nomark \\

PhysGym~\citep{chen2025physgymbenchmarkingllmsinteractive}
    & \fullmark
    & \nomark
    & \fullmark
    & \partmark
    & \partmark
    & \nomark \\

AdaPlanBench~\citep{liu2026adaplanbenchevaluatingadaptiveplanning}
    & \partmark
    & \partmark
    & \nomark
    & \nomark
    & \fullmark
    & \fullmark \\

PlanBench-XL~\citep{liu2026planbench}
    & \partmark
    & \partmark
    & \nomark
    & \nomark
    & \nomark
    & \fullmark \\


\midrule

\textbf{\bench{} (Ours)}
    & \fullmark
    & \fullmark
    & \fullmark
    & \fullmark
    & \fullmark
    & \fullmark \\

\bottomrule
\end{tabular}%
}

\caption{
Positioning of \bench{}. For each work, the table reports whether each trait is fully (\fullmark), partially (\partmark) or not (\nomark) addressed. Detailed explanations of each terminology are provided in Appendix \ref{app:positioning_terms}.
}
\vspace{-0.5em}
\label{tab:positioning}
\end{table*}

However, as summarized in Table~\ref{tab:positioning}, existing benchmarks do not fully capture these capabilities.
The evidence are often given \citep{hua2025inductionbench,zheng2026herosjourney} or leaked from environment \citep{liu2026adaplanbenchevaluatingadaptiveplanning}, and the environments are often stationary \citep{he2025idea,zheng2026newtonbench,chen2026causalgamebenchmarkingcausalthinking}.
These limitations motivate our central question: \textbf{Can LLM agents proactively explore to understand and adapt to the environmental dynamics?}
Answering the question needs a benchmark with: 
\textbf{(1) Self-directed environment exploration},
where agents must actively gather evidence to uncover environmental rules, rather than relying on provided evidence or predefined task-completion objectives;
and \textbf{(2) Unannounced environmental changes},
where latent rules may shift without explicit notification, requiring agents to recognize when previously acquired knowledge becomes outdated and re-explore to adapt.

To this end, we introduce \textbf{\bench{}}, a benchmark for proactive exploration and test-time adaptation in evolving environments.
In this work, we instantiate environmental dynamics as latent rules,
a mapping from agents' action to its outcome \citep{ICLR2025_b782a346,ICLR2024_a9342824}.
Understanding the rules corresponds to learning the environment, while a change in the rule represents evolving environments.
\bench{} consists of 256 evaluation and 1000 training instances over four rule families. 
In the default setting, the agent decides itself on when to interact, which input values to query, and how many evidence to collect.
Meanwhile, unannounced \dynamic{} occurs during interaction, where agents must detect from altered exploration outcome.
To ensure that even agents with limited exploration effort have opportunities to recognize \dynamic{}, evaluation cases are presented sequentially in small batches.
Agents can thus detect \dynamic{} through either further exploration or performance degradation.
This protocol also yields a performance curve over the interaction, enabling in-process evaluation for further analysis.

We evaluate eight leading LLMs on \bench{}.
Our results suggest that proactive exploration and test-time adaptation remains highly challenging.
While all models get below 70\% calibrated accuracy in default setting
, giving extra evidence or removing rule changes improves average performance by 9.4\% and 6.2\% significantly, suggesting that models are considerably better at exploiting given evidence in static environment.
We further find that exploration quality is more important than quantity, and agents often report partially correct but incomplete rules and rule changes.
Thess findings show that current LLM agents still lag in proactive exploration and test-time adaptation in evolving environment.
We summarize our main contributions as:
\begin{itemize}
    \item \textbf{Scalable Construction Pipeline:} We introduce \bench{}, a scalable LLM-based generation pipeline to synthesize complex and diverse evaluation instances without extra human labor.
    \item \textbf{Evolving Environment Design:} We instantiate proactive and adaptive exploration in an evolving environment governed by latent rules and rule changes, reflecting real-world challenges and providing a testbed for future training.
    \item \textbf{Thorough Analysis:} We evaluate eight LLMs and reveal that current agents remain limited in proactive exploration and dynamic adaptation. 
\end{itemize}
Together, our work provides a foundation for understanding the limitations of current LLM agents and serves as a training playground for improving their exploration and adaptation capabilities.
\vspace{-0.1em}
\begin{figure*}[t]
    \centering
    \includegraphics[width=\linewidth]{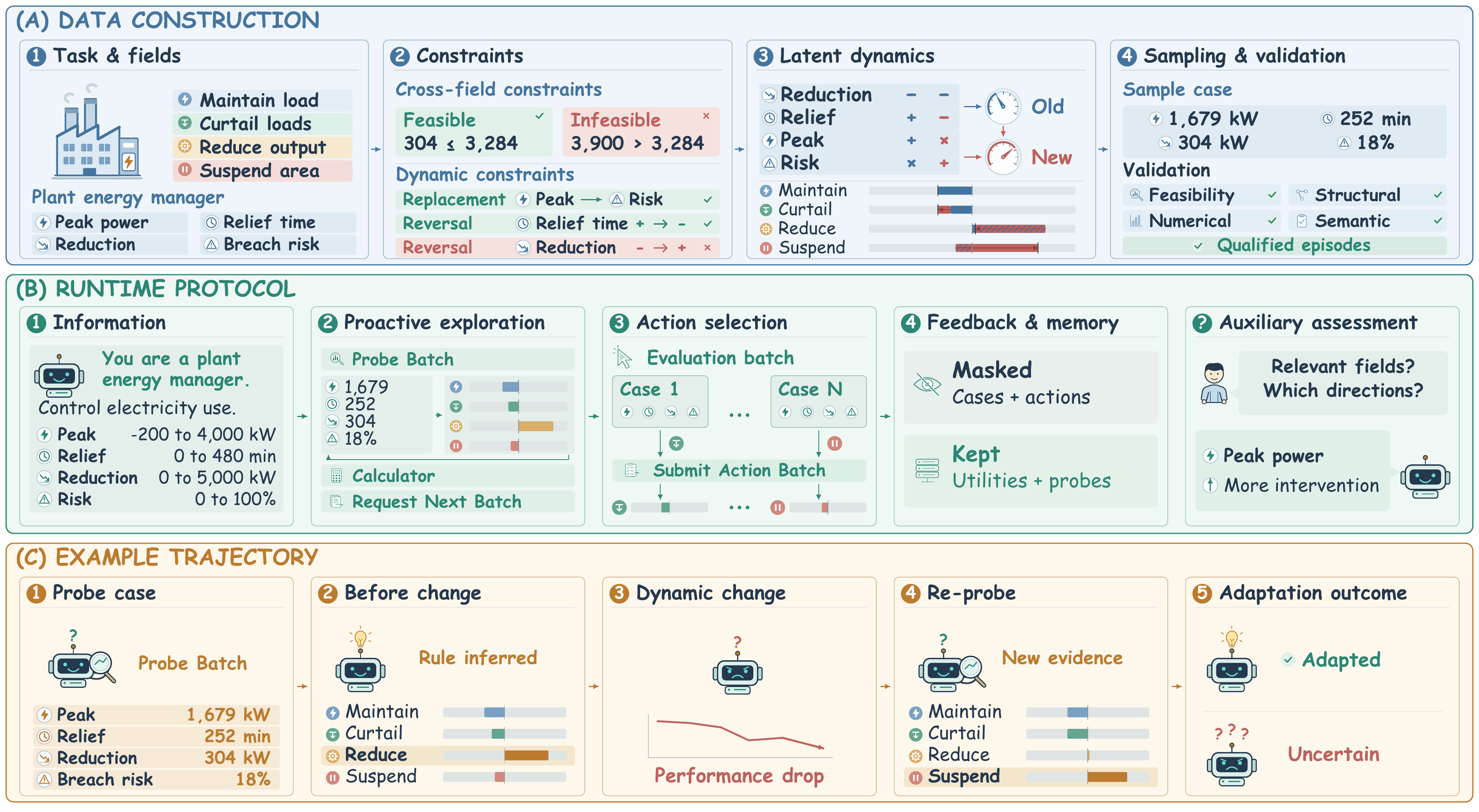}
    \caption{
        Overview of \bench{}. \textit{Data construction} shows the generation pipeline. \textit{Runtime protocol} shows how agents explore and how are they assessed. \textit{Example Trajectory} shows a complete example of an episode.
    }
    \vspace{-0.5em}
    \label{fig:overview}
\end{figure*}

\section{Related Work}

\paragraph{Proactive Exploration for Environment Learning.}
Recent work moves beyond reasoning over given evidence by letting agents proactively explore to learn the environment.
Prior work studies proactive learning through scientific discovery~\citep{zheng2026newtonbench}, interactive learning~\citep{he2025idea}, hypothesis testing~\citep{banatt2024wilt}, long-horizon induction~\citep{xu2026odysseyarena}, and alien-world exploration~\citep{zhang2026explorationbenchmeasuringaisystems}.
These works test proactive exploration and targeted at environment learning, but the environment typically remains static, and sometimes leaks information not queried by the agent \citep{he2025idea}.
In contrast, \bench{} prevents information leakage and introduces unannounced \dynamic{}, requiring agents to re-explore and adapt accordingly.


\paragraph{Test-time Adaptation.}
A complementary line of work studies adaptation when task-relevant conditions change during interaction.
Related work studies adaptation to changing conditions through concept drift~\citep{gama2014survey}, deployment-time dynamics grounding~\citep{chen2026testtime}, progressively revealed constraints~\citep{liu2026adaplanbenchevaluatingadaptiveplanning}, unreliable tools~\citep{liu2026planbench}, and changing tool costs~\citep{liu2026costbenchevaluatingmultiturncostoptimal}.
However, these works emphasize statistical adaptation or task-oriented replanning, whereas \bench{} requires agents to recognize unannounced \dynamic{}, revise their hypotheses, and proactively re-explore.
More related works are provided in Appendix~\ref{app:related_work}.

\section{Benchmark Construction}

\subsection{Data Generation}

\paragraph{Definitions.}
We consider decision-making environments where an agent selects actions based on observable input fields.
Each \textit{case} contains a set of field values, and the agent chooses from four actions $\mathcal{A}=\{a_0,a_1,a_2,a_3\}$ ordered by intervention strength.
A subset of fields, termed \textit{factors}, determines the appropriate intervention, with each factor having a positive or negative directional effect.
The latent environmental dynamics map each case to an intervention score $s\in[-1,1]$, where higher scores favor stronger interventions.
A fixed utility function maps this score to action utilities, rewarding actions whose intervention levels better match the score.
An \textit{episode} is a complete evaluation instance comprising a task and a sequence of cases governed by potentially changing environmental dynamics.

\paragraph{Task Specification.}
Each task defines an operational scenario, an agent role, and four ordered actions based on predefined blueprints across finance, healthcare, cybersecurity, and manufacturing.
We use LLM-assisted generation to construct a field catalog with semantic descriptions, numerical ranges, and sampling distributions.
Each episode selects relevant factors and irrelevant distractors from the catalog.
We further impose (1) \textit{cross-field constraints} to ensure feasible field combinations and (2) \textit{dynamics constraints} to specify admissible factor interactions and environmental changes.

\paragraph{Environmental Dynamics.}
We instantiate four families representing different decision-making strategies~\citep{einhorn1970nonlinear,payne1993adaptive}.
\textit{Additive} dynamics aggregate evidence across factors, \textit{conjunctive} and \textit{disjunctive} dynamics require all or any conditions to be satisfied, and \textit{lexicographic} dynamics follow a factor priority order.
The resulting evidence is calibrated to a common intervention score.
In static settings, the underlying dynamics remain fixed, while dynamic settings introduce an unannounced \dynamic{} that replaces one relevant factor and reverses another factor's direction within the same family.
These changes are constrained to be semantically plausible and induce meaningful yet moderate shifts in optimal actions.
Detailed formalizations are provided in Appendix~\ref{app:task_setup_details}.

\paragraph{Episode Generation and Validation.}
We construct episodes by instantiating latent dynamics and sampling feasible evaluation cases using established field catalog and following the constraints.
Candidates undergo structural, numerical, and semantic validation to ensure valid configurations, non-degenerate action distributions, and plausible environmental changes.
A further LLM-based review evaluates field realism, change plausibility, and distractor quality.
We retain only qualified episodes from candidates for the final benchmark.
Detailed procedures and prompts are provided in Appendix~\ref{app:data}.
To ensure high quality, human study of the generated instances are given in Appendix~\ref{app:human-study}.

\subsection{Runtime Protocol}
\label{sec:runtime_protocol}

\paragraph{Interaction Protocol.}
At the beginning of each episode, the agent receives the task description, input fields and ranges, available actions, definitions of the four dynamics families, and interaction tools.
The instantiated family, relevant factors, parameters, and occurrence and timing of environmental changes remain hidden.
Each episode proceeds through evaluation batches preceded by exploration windows.
The agent may use \textit{Probe Batch} to query self-designed cases and observe utilities for all actions, or \textit{Request Next Batch} to proceed without probing.
For each revealed evaluation batch, \textit{Submit Action Batch} collects action choices and returns utility feedback.
A \textit{Calculator} supports numerical reasoning.
The agent decides whether, what, and how much to explore, while receiving cumulative utility feedback throughout the episode.

\paragraph{Evaluation History Masking.}
By default, completed evaluation cases and selected actions are removed from subsequent context, while their realized utilities remain available.
In contrast, self-collected probe observations are retained.
This isolates proactive exploration from environment-provided evidence while preserving non-diagnostic feedback that may signal environmental changes.

\paragraph{Auxiliary Assessment.}
We conduct necessary assessments to directly evaluate whether agents identify relevant factors, their directional effects, and changes in environmental dynamics.
Assessments are collected immediately before \dynamic{} and at episode completion in dynamic settings, and only at completion in static settings.
They are excluded from subsequent interaction context to avoid influencing agent behavior.
Please refer to  Appendix~\ref{app:runtime_settings} for detailed and additional runtime settings.
\begin{table*}[t]
\centering
\resizebox{\textwidth}{!}{
\small
\begin{tabular}{l cc cc cc}
\toprule
\multirow{2}{*}{\textbf{Model Name}} & \multicolumn{2}{c}{\textbf{End-to-end Outcome}} &
\multicolumn{2}{c}{\textbf{Exploration Behavior}} &
\multicolumn{2}{c}{\textbf{Adaptation Performance}} \\
\cmidrule(lr){2-3} \cmidrule(lr){4-5} \cmidrule(lr){6-7}
& EA (\%) $\uparrow$ & CA (\%) $\uparrow$ & Mean Uni. Probes & Mean Re-probes & Factor Rec. (\%) $\uparrow$ & Change Rec.
(\%) $\uparrow$
\\
\midrule
\textit{Llama-3.1-8B}          & 30.67              & 8.13               & 4.77   & 0.24  & 52.54              & 6.30
\\
\textit{Qwen3.5-9B}            & 32.17              & 12.31              & 7.81   & 1.04  & 39.40              & 13.80
\\
\textit{Qwen3.8-27B}           & 45.95              & 37.73              & 42.20  & 5.46  & 65.87              & 25.93
\\
\textit{Gemma-4-12B}           & 37.98              & 20.59              & 12.52  & 0.13  & 53.21              & 10.34
\\
\textit{DeepSeek-V4.1-Flash}   & 58.01              & 52.87              & 113.42 & 16.04 & \underline{78.07}  & 44.52
\\
\textit{GPT-6-Luna}            & 46.50              & 35.10              & 22.84  & 0.73  & 56.66              & 9.96
\\
\textit{GPT-6-Sol}             & \underline{65.69}  & \underline{61.87}  & 99.03  & 6.95  & 73.79              &
\underline{47.56}
\\
\textit{Gemini-3.8-Flash}      & \textbf{69.47}     & \textbf{67.25}     & 88.12  & 10.25 & \textbf{78.30}     &
\textbf{63.68}
\\
\bottomrule
\end{tabular}
}
\caption{Main evaluation results in the \emph{default} setting. Scores in \textbf{bold} indicate the
best performance among all models, and \underline{underlined} scores denote the second-best performance. 
\emph{Exploration Behavior} metrics omit arrows and highlights, as neither direction is inherently better.
Metrics marked with (\%) are reported in percentage form.}
\vspace{-0.5em}
\label{tab:main_results}
\end{table*}

\section{Experiments}

\subsection{Settings}
\label{sec:settings}

\paragraph{Models.}
Our evaluation includes proprietary and leading open-source models with distinct parameter scale.
Proprietary models include \emph{GPT-6-Luna}, \emph{GPT-6-Sol}~\citep{openai2026gpt6}, and \emph{Gemini-3.8-Flash}~\citep{deepmind2026gemini38flash}, while open-source models include \emph{DeepSeek-V4.1-Flash}~\citep{deepseekai2026v41flash}, \emph{Gemma-4-12B}~\citep{gemmateam2026gemma4}, \emph{Llama-3.1-8B}~\citep{grattafiori2024llama3}, \emph{Qwen3.5-9B}~\citep{qwent2026qwen35}, and \emph{Qwen3.8-27B}~\citep{qwent2026qwen38}.
Detailed reasoning and inference configs are specified in Appendix~\ref{app:model_setup}.

\paragraph{Experimental Settings.}
Each setting is a combination of configurations of (1) Whether \dynamic{} occur. (2) Whether evaluation batches are masked. (3) Whether probings are mandatory.
We evaluate agents under the following settings:

\noindent \textbf{(1) Default.}
\dynamic{} occurs during interactions. Evaluation batches are masked. Probing amount is self-decided by the agent.

\noindent \textbf{(2) Stationary.}
The same as \textit{default} except that \dynamic{} does not occur during interactions.

\noindent \textbf{(3) Evidence Given.}
The same settings as \textit{default} except that Evaluation batches are not masked.

\noindent \textbf{(4) Forced-Probe.}
The same settings as \textit{default} except that probings are mandatory in every window with a minimal number of cases required.

\noindent We evaluate the main results under \textit{default} setting and the other settings are for analysis.

\noindent \subsection{Metrics}
We evaluate end-to-end performance, exploration behavior and adaptation performance.
The detailed metrics definitions are provided in Appendix~\ref{app:metric_definitions}.
Here we give a brief explanation of each one.

\noindent \textbf{(1) Exact Accuracy (EA)}
measures the strict correctness of the agent's decisions.
Each evaluation case receives credit only when the agent selects the optimal action with largest utility feedback.
We report the mean credit over evaluation cases.

\noindent \textbf{(2) Calibrated Accuracy (CA)}
is the calibrated version of EA, where near-optimal choices are granted partial credit based on the actual utility received.
The score is further calibrated so that a uniformly random policy obtains zero in expectation and the optimal policy obtains full score.

\noindent \textbf{(3) Mean Unique Probes}
measures the amount of exploration.
We count the average number of distinct probes queried by the agent.
Identical probes issued before and after \dynamic{} are treated as distinct, since they may help recognize \dynamic{}.

\noindent \textbf{(4) Mean Re-probes after Change} measures the agent's verification effort after \dynamic{}.
We count the distinct pre-change probe points that the agent queries again later.
It is the exact complement of Mean Unique Probes on post-change probes.

\noindent \textbf{(5) Factor Recognition}
measures how accurately the agent identifies the factors of the rule through the auxiliary assessment.
We compare the predicted factor set with the ground-truth set by F1, and averaging the pre- and post-change values.

\noindent \textbf{(6) Change Recognition}
measures whether the agent correctly identifies \emph{what changed} in the rule.
We represent \dynamic{} as a set of atomic changes and score the changes inferred from the agent's pre- and post-change assessments by set F1.

\subsection{Results}
\label{sec:results}

\begin{takeaway}
    Proactive and adaptive exploration remains highly challenging for most models.
\end{takeaway}

\paragraph{All models fall below 70\% calibrated accuracy, and weaker models select more costly actions.}
Table~\ref{tab:main_results} shows a clear performance divide across models.
\textit{Gemini-3.8-Flash} achieves the strongest end-to-end performance, but still score below 70\%.
Moreover, the gap between EA and CA widens substantially as model performance decreases. 
It ranges from about 2 to 5 points for the top three models, but grows to 8 to 22 points for the remaining models.
This suggests that weaker agents not only make incorrect decisions more frequently, but their mistakes also tend to incur larger utility losses.
The results highlight the difficulty of proactive and adaptive exploration under evolving environments.

\begin{takeaway}
    Greater exploration and verification effort do not necessarily translate into better end-to-end performance across models.
\end{takeaway}

\paragraph{Exploration effort yields highly unequal returns across models.}
Table~\ref{tab:main_results} reveals variation in exploration behavior, with \textit{DeepSeek-V4.1-Flash} issuing approximately 24 times as many probes as \textit{Llama-3.1-8B}.
However, this variation does not translate into proportional performance gains.
Despite probing the most, \textit{DeepSeek-V4.1-Flash} achieves lower CA than both \textit{Gemini-3.8-Flash} and \textit{GPT-6-Sol}.
Similarly, \textit{Gemini-3.8-Flash} outperforms \textit{GPT-6-Sol} with fewer probes.
These contrasts reveal a discrepancy between exploration volume and realized performance, as models with fewer interactions can achieve substantially better outcomes.

\paragraph{Additional verification does not close the performance gap.}
This discrepancy extends to post-change exploration.
\textit{DeepSeek-V4.1-Flash} performs considerably more re-probes than the two strongest models, yet still achieves lower end-to-end performance.
Thus, additional opportunities to verify previously learned relationships do not lead to more successful adaptation.
Together, these results highlight the limitations of measuring exploration through interaction counts alone and underscore the importance of evaluating how effectively agents acquire and exploit informative evidence.

\begin{takeaway}
    Adaptation ability distinguishes effective agents more clearly than exploration efforts.
\end{takeaway}

\begin{figure*}[t]
  \centering
  \includegraphics[width=\textwidth]{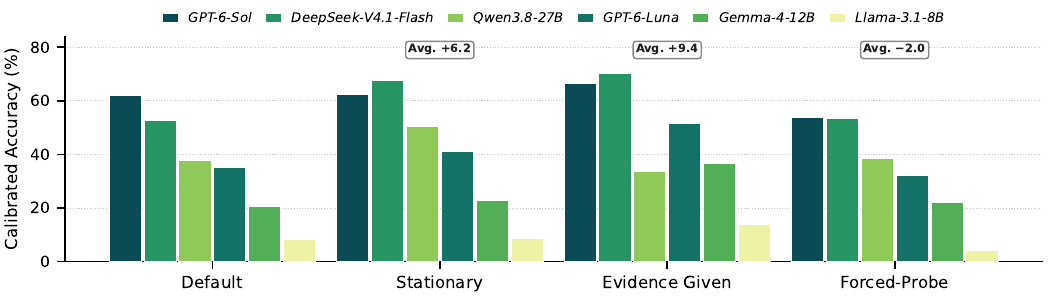}
  \caption{Calibrated Accuracy under \textit{default} and the three other settings. Boxed annotations give the average CA gain over Default across all models. The full results and significance analysis provided in Appendix~\ref{app:full_analysis} and~\ref{app:significance_analysis}.}
  \label{fig:diagnostics}
\end{figure*}

\paragraph{Recognition is a stronger indicator of end-to-end performance.}
Table~\ref{tab:main_results} shows that differences in recognition ability are more strongly reflected in final performance than differences in exploration effort.
The four strongest models by CA follow exactly the same ranking in Change Recognition, while the remaining models achieve at most 13.80\% Change Recognition and substantially lower CA.
Factor Recognition exhibits a similar, though relatively less pronounced, pattern.
In contrast, exploration effort does not distinguish model performance as consistently.
Further detailed correlation analysis is provided in Appendix~\ref{app:correlation_analysis}.
\section{Analysis}
\label{sec:analysis}

Section~\ref{sec:results} reports models' behaviors under \textit{default} setting.
Here we ask why they fail by ablations.

\begin{itemize}[leftmargin=1.2em,itemsep=1pt]
  \item \textbf{\Dynamic{} Adaptation.} How much of the performance gap does \dynamic{} cause?
  \item \textbf{Evidence Sufficiency.} How much is caused by the burden of acquiring evidence proactively?
  \item \textbf{Exploration Quantity vs. Quality.} Can forcing agents to explore more close the gap?
\end{itemize}

\noindent
Figure~\ref{fig:diagnostics} reports Calibrated Accuracy (CA) for Default
and the three diagnostic settings.

\begin{takeaway}
Removing unannounced \dynamic{} significantly recovers agents' end-to-end outcome.
\end{takeaway}

\paragraph{Removing \dynamic{} recovers performance significantly for most models.}
The Stationary setting only removes \dynamic{} compared with \textit{default} setting.
It raises CA by up to 14.7\% for \textit{DeepSeek-V4.1-Flash} in
Figure~\ref{fig:diagnostics}, and 6.2\% on average.
Though \textit{GPT-6-Sol} barely changes, it is because its \textit{default} performance is already near its stationary ceiling, consistent with its second-best Change Recognition in Table~\ref{tab:main_results}.
The weakest models improves by less 1\%, so removing \dynamic{} barely help agents that cannot even learn in \textit{stationary} setting.
\Dynamic{} is thus a first-order bottleneck for models with basic learning competence.

\begin{takeaway}
Agents exploit effectively when sufficient evidence is provided externally.
\end{takeaway}

\paragraph{Most models improve substantially when sufficient evidence is given.}
The \textit{Evidence Given} setting let passive evidence accumulate across batches.
Six of the seven models improve, four by more than 15 CA points. 
The resulting gains identify proactive evidence acquisition as a key bottleneck: models learn better from available evidence than they acquire it on their own.
\textit{Qwen3.8-27B} is the only model that degrades under Evidence Given, but it also cuts exploration effort the most as shown in Appendix~\ref{app:analysis_results}.
This shows that sufficient given evidence that become \emph{stale} later can mislead models that trust them without re-verifying, consistent with the near-zero re-probing effort in this setting.

\begin{figure*}[t]
  \centering
  \includegraphics[width=\textwidth]{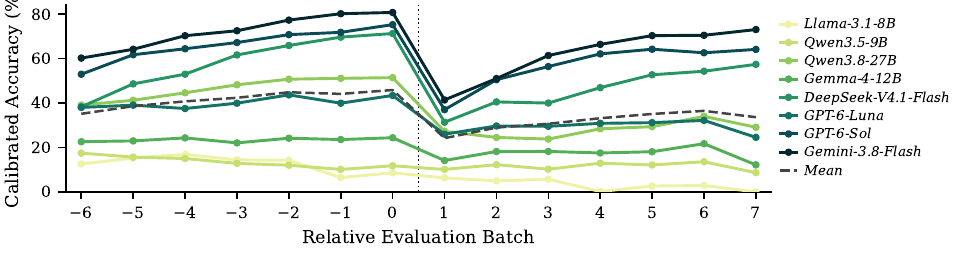}
  \caption{Calibrated Accuracy per evaluation batch under \textit{default}, with batch 0 marking the last pre-change batch.}
  \vspace{-0.5em}
  \label{fig:ca-curve}
\end{figure*}

\begin{takeaway}
Effective exploration depends more on explored evidence quality than quantity.
\end{takeaway}

\paragraph{Forcing agents to explore more barely help.}
If the gap reflected exploration willingness or amount, Forced-Probe should close it.
In \textit{forced-probe} setting, probing quantity indeed shifts drastically.
As shown in Appendix~\ref{app:analysis_results}, Mean unique probes rise from 33.9 to 53.8 across all models, and the most reluctant explorers scale up their efforts by two to five times.
Re-probes after the rule change also rise substantially.
Yet performance stays flat or even degrades, with a mean CA change of $-2.0\%$.

\paragraph{Probe informativeness emerges as a likely bottleneck.}
Poor performance may stem from insufficient exploration effort, weak induction from available evidence, or uninformative probe selection.
However, Forced-Probe weakens the first explanation, as increasing exploration effort fails to improve performance, while Evidence Given weakens the second, as models improve substantially when sufficient evidence is provided.
Together, these findings suggest that the inability to select informative probe cases is a key remaining bottleneck, although probe quality is not directly measured.

\paragraph{Synthesis.}
Agents learn well from given evidence, but they neither acquire comparably useful evidence on their own, nor detect and revise around \dynamic{}.
The binding constraints on current LLM agents in evolving environments are proactive exploration \emph{quality} and test-time adaptation, not exploration quantity or basic rule learning.
More ablation studies are provided in Appendix~\ref{app:ablation}.
\section{Error Analysis}
\label{sec:error-analysis}

Section~\ref{sec:analysis} showed why the agents fail. We now look inside the main results and ask what the failure looks like.
Specifically, we examine:

\begin{itemize}
  \item When does accuracy drop and whether the agent recovers its performance later?
  \item Does re-probing helps recognizing \dynamic{}?
  \item How factor and change recognition fails?
\end{itemize}

\noindent Results in Figure~\ref{fig:ca-curve}, Table~\ref{tab:reprobe-correlation} and Table~\ref{tab:recognition-pr} give comprehensive analysis for answering the questions.

\begin{takeaway}
\Dynamic{} causes an immediate performance drop, and no model recovers at last.
\end{takeaway}

\paragraph{Accuracy drops immediately at \dynamic{} and never fully recovers.}
Figure~\ref{fig:ca-curve} aligns each episode at its \dynamic{}.
Every model that climbs above the random baseline loses a large share of that progress at the first post-change batch, with an average loss of 26.9 points across the six models that operate above the floor.
Each of these drops is significant, and recovery is partial.
Averaging over the final two batches, all six models remain 6.9 to 19.2 points below their own pre-change plateau.
Weak models stay near the random baseline even before \dynamic{}, so there is little left to lose and no significant drop. 
\textit{Llama-3.1-8B} further decrease to 1.3\% by the final batch, possibly a performance decay under long context.
The rule change is therefore not a transient disturbance but has a huge impact on the subsequent trajectory.

\begin{table}[t]
  \centering
  \small
  \resizebox{\columnwidth}{!}{%
  \begin{tabular}{lcccc}
    \toprule
    & & & \multicolumn{2}{c}{Spearman $\rho$} \\
    \cmidrule(lr){4-5}
    Model & Re-probes & Change Rec.(\%) & count & rate \\
    \midrule
    Llama-3.1-8B & 0.24 & 6.30 & 0.05 & 0.05 \\
    Qwen3.5-9B & 1.04 & 13.80 & 0.01 & 0.03 \\
    Qwen3.8-27B & 5.46 & 25.93 & \textbf{0.66} & \textbf{0.64} \\
    Gemma-4-12B & 0.13 & 10.34 & 0.05 & 0.05 \\
    DeepSeek-V4.1-Flash & 16.04 & 44.52 & \textbf{0.25} & \textbf{0.30} \\
    GPT-6-Luna & 0.73 & 9.96 & \textbf{0.30} & \textbf{0.30} \\
    GPT-6-Sol & 6.95 & 47.56 & \textbf{0.59} & \textbf{0.52} \\
    Gemini-3.8-Flash & 10.25 & 63.68 & \textbf{0.32} & \textbf{0.39} \\
    \midrule
    \multicolumn{3}{l}{Within model} & \textbf{0.37} & \textbf{0.28} \\
    \multicolumn{3}{l}{Across models} & \underline{0.86} & 0.45 \\
    \bottomrule
  \end{tabular}}
  \caption{Correlation between Mean Re-probes and Change Recognition on \textit{default}, where rate normalizes re-probes by distinct pre-change probe number. Bold and underlined values indicate $p<0.001$ and $p<0.01$.}
  \vspace{-0.5em}
  \label{tab:reprobe-correlation}
\end{table}

\begin{takeaway}
Re-probing provides additional evidence, but successful adaptation also depends on how effectively a model exploit evidence.
\end{takeaway}

\paragraph{Re-probing supports within-model change recognition.}
Table~\ref{tab:reprobe-correlation} shows significant positive correlations between re-probing and Change Recognition for all models performing above the recognition floor.
These correlations persist after normalizing re-probes by pre-change exploration and when pooling model-centered episodes.
This suggests that revisiting previously explored points provides direct evidence which help agents detect inconsistencies with their existing hypotheses and identify changes in environmental dynamics.
The consistent within-model associations highlight the value of targeted verification during adaptation.

\paragraph{Re-probing alone does not explain cross-model differences.}
Although raw re-probe counts correlate with Change Recognition across models, this relationship becomes insignificant after normalization by pre-change exploration.
This suggests that the apparent association is partly driven by differences in exploration volume, rather than verification effectiveness alone.
For instance, \textit{DeepSeek-V4.1-Flash} performs more re-probes than \textit{Gemini-3.8-Flash}, yet achieves lower Change Recognition.
One possible explanation is that models differ in their ability to exploit the collected evidence, particularly in interpreting before-and-after observations and updating their hypotheses.
Thus, effective change recognition depends not only on the availability of verification evidence but also on the agent's ability to reason over it.

\begin{table}[t]
  \centering
  \small
  \resizebox{\columnwidth}{!}{%
  \begin{tabular}{lcccc}
    \toprule
    & \multicolumn{2}{c}{Factor Rec.} & \multicolumn{2}{c}{Change Rec.} \\
    \cmidrule(lr){2-3}\cmidrule(lr){4-5}
    Model & P & R & P & R \\
    \midrule
    Llama-3.1-8B & 38.2 & \textbf{84.5} & \textbf{10.5} & 6.5 \\
    Qwen3.5-9B & \textbf{43.9} & 35.9 & \textbf{19.2} & 14.9 \\
    Qwen3.8-27B & \textbf{81.1} & 55.6 & \textbf{66.9} & 23.1 \\
    Gemma-4-12B & 52.8 & \textbf{53.5} & \textbf{18.6} & 10.5 \\
    DeepSeek-V4.1-Flash & \textbf{86.9} & 70.8 & \textbf{77.8} & 39.3 \\
    GPT-6-Luna & \textbf{63.8} & 50.7 & \textbf{31.3} & 9.2 \\
    GPT-6-Sol & \textbf{79.1} & 69.1 & \textbf{63.6} & 45.6 \\
    Gemini-3.8-Flash & \textbf{94.9} & 66.9 & \textbf{83.1} & 58.3 \\
    \midrule
    Mean & \textbf{67.6} & 60.9 & \textbf{46.4} & 25.9 \\
    \bottomrule
  \end{tabular}}
  \caption{Precision and recall for Factor Rec. and Change
  Rec. on \textit{default}. The larger value is \textbf{bolded}.
  }
  \vspace{-0.5em}
  \label{tab:recognition-pr}
\end{table}

\begin{takeaway}
Agents tend to settle on incomplete representations of environmental dynamics.
\end{takeaway}

\paragraph{Agents prematurely settle on incomplete environmental models.}
Table~\ref{tab:recognition-pr} shows that precision generally exceeds recall in both Factor and Change Recognition, suggesting that agents recover mostly correct but incomplete representations of environmental dynamics.
This pattern points to a potential failure of hypothesis refinement, where agents may settle on plausible explanations without sufficiently exploring alternative dependencies.
Such incomplete representations can further hinder adaptation, as changes involving overlooked factors may remain undetected.
Indeed, in 36.3\% of eligible episodes, agents report identical pre- and post-change assessments despite changes in the environment.
This failure is particularly pronounced for \textit{GPT-6-Luna} and \textit{Qwen3.8-27B}, affecting 65.6\% and 54.2\% of their episodes, respectively.
Detailed episode-level results are provided in Appendix~\ref{app:episode-error-analysis}.

\begin{takeaway}
Agents struggle to recognize changes in how existing factors influence outcomes.
\end{takeaway}

\paragraph{Agents struggle to revise previously learned factors.}
All evaluated models recognize factor replacements more reliably than direction reversals.
At the episode level, agents identify both replacements but miss the reversal in 17.7\% of episodes, compared with only 4.1\% for the converse.
This suggests that agents are better at updating which factors matter than revising how existing factors affect outcomes.
Unlike replacements, reversals require agents to correctly infer directional effects both before and after \dynamic{}, potentially making previous knowledge harder to revise.
This highlights a limitation in hypothesis revision, where agents may retain outdated interpretations despite changes in environmental dynamics.
Detailed results are provided in Appendix~\ref{app:episode-error-analysis}.

\paragraph{Synthesis.}
Together, these findings suggest that a central obstacle to adaptation is the persistence of unchallenged assumptions.
Agents may form plausible but incomplete environmental models and fail to reconsider established rules as the environment evolves.
This creates a potential blind spot where insufficient exploration and strong hypothesis persistence reinforce each other.
A detailed case study is given in Appendix~\ref{app:case_study} for more clarity.
\section{Conclusion}

In this work, we introduce \textbf{\bench{}}, a benchmark for evaluating proactive exploration and test-time adaptation of LLM agents in evolving environments. 
By instantiating environmental dynamics as latent rules, we requires agents to decide when, what, and how much to explore, while recognizing \dynamic{} and revising their hypothesis accordingly.
Our results show that current LLM agents remain limited in both \textbf{proactive exploration} and \textbf{test-time adaptation}. 
These findings highlight a broader challenge in enabling agents to autonomously acquire evidence while keeping their understanding of evolving environments up to date.
Addressing this challenge may require agents to actively challenge established assumptions and revise their knowledge as the environment changes.
We envision \bench{} as a foundation for developing agents capable of reliable exploration and adaptation in dynamic environments.

\section*{Limitations}

\paragraph{Simplified Environmental Dynamics.}
We model environmental dynamics as latent rules and \dynamic{}, providing a controlled but necessarily simplified abstraction of real-world environments.
While this formulation cannot fully capture real-world complexity and uncertainty, agents already struggle with proactive exploration and adaptation even in these simplified settings.

\paragraph{Indirect Assessment of Exploration Quality.}
Our exploration metrics primarily measure effort rather than quality.
We omit direct quality metrics based on spatial diversity or decision-region coverage, as they inherently favor extensive probing and yield incomparable scores across heterogeneous rule families.
Instead, we use corresponding ablations settings to indirectly examine whether ineffective evidence acquisition limits performance.

\paragraph{Limited Coverage of Environmental Changes.}
Our evaluation currently focuses on within-family \dynamic{} involving factor replacement and direction reversal.
Due to time and computational resource constraints, we leave cross-family changes, alternative change types, and systematically varying change magnitudes for future work.

\section*{Ethics Statement}

\paragraph{Offensive Content.}
\bench{} consists of synthetically generated scenarios across domains.
The evaluation instances are constructed without collecting sensitive user data. 
The benchmark is designed solely for research evaluation and does not involve offensive content.

\paragraph{Licenses.}
Our code will be released under the MIT license, and \bench{} will be distributed under a Creative Commons license to support academic research. We will ensure that our use of third-party models and tools complies with their respective licenses and terms of use.

\paragraph{Models.}
All open-source models were hosted and executed locally using the vLLM library \citep{kwon2023efficient}, while all closed-source models were accessed through their official APIs.
The experimental settings are detailed in~\ref{app:model_setup} for reproducibility.

\paragraph{Data Annotations.}
All data annotation was performed by PhD-level researchers with relevant expertise, ensuring that the process was conducted responsibly and following ethical standards.

\bibliography{custom}

\appendix

\section{Terms Explanation in Positioning Table}
\label{app:positioning_terms}
We define each trait in Table~\ref{tab:positioning} below.

\begin{itemize}
    \item \textbf{Proactive Exploration.} The agent decides itself on when, what, and how much to interact for collecting evidence. Evidence should be mostly self-queried, instead of given or leaked from the environment.

    \item \textbf{Adaptive Exploration.} The environment dynamics change during interaction without notifications, and the agent must detect the shift and explore again.

    \item \textbf{Env-learning Exploration.} Exploration aims to learn how the environment works, rather than to complete a predefined task.

    \item \textbf{In-Process Evaluation.} Performance is measured throughout exploration instead of only at the end, yielding a performance curve over the whole interaction process.

    \item \textbf{Heterogeneous Rules.} Latent rules span structurally and cognitively distinct families rather than variations of similar forms.

    \item \textbf{Scalable Generation.} evaluation instances are procedurally generated and the benchmark size can be scaled up easily without human labor.
\end{itemize}

\begin{figure}[t]
    \centering
    \includegraphics[width=\linewidth]{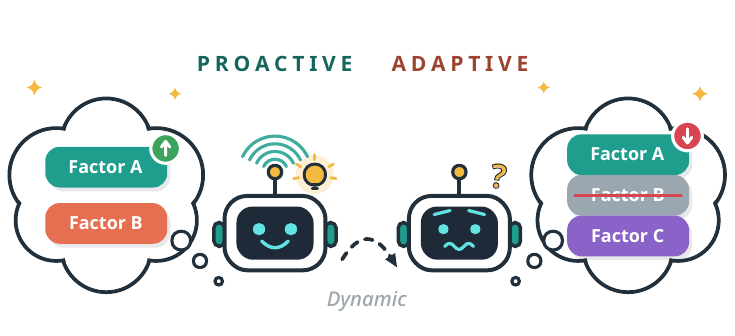}
    \caption{
        Motivation of \bench{}. An agent proactively explore and get confused when rule changes.
    }
    \vspace{-0.5em}
    \label{fig:motivation}
\end{figure}

\section{More on Related Work}
\label{app:related_work}
\paragraph{Passive Rule Learning.}
Prior work has examined whether LLMs can infer latent rules from given evidence.
InductionBench evaluates inductive reasoning by requiring models to recover underlying transformation rules from a fixed set of observations \citep{hua2025inductionbench}.
Patterns Over Principles studies the robustness of rule induction under noisy observations, revealing the fragility of LLM rule abstraction \citep{li2025patterns}.
HERO'S JOURNEY extends rule induction to rule executions, where agents not only infer hidden rules from demonstrations but also apply them in real tasks \citep{zheng2026herosjourney}.
However, these settings are largely passive, where the evidence used is predetermined.
In contrast, \bench{} embeds rule learning in an interactive decision process where agents can actively construct probes to gather diagnostic evidence before making decisions.

\section{Environment Formalization}
\label{app:task_setup_details}

\begin{figure}[t]
    \centering
    \includegraphics[width=\columnwidth]{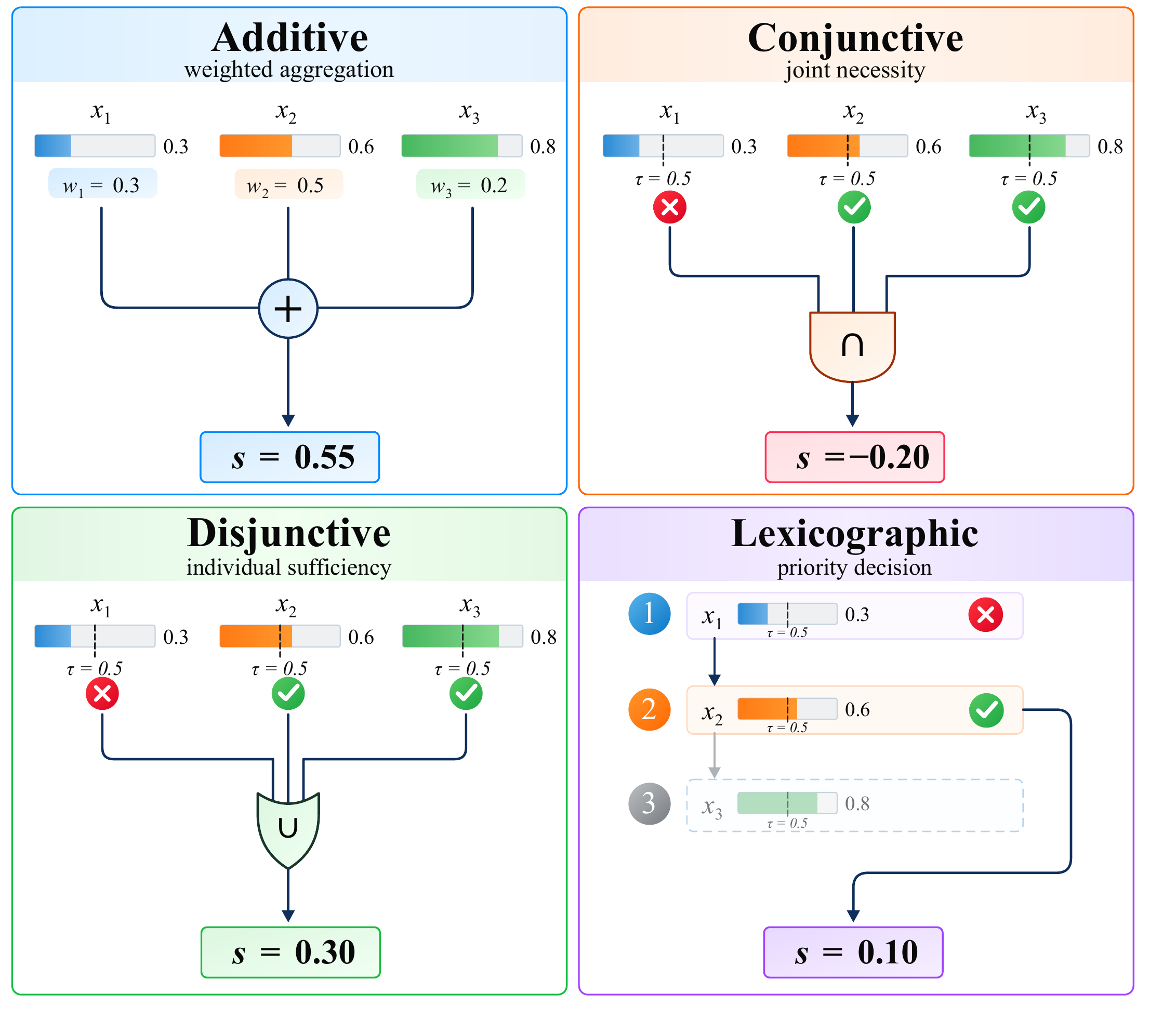}
    \caption{Illustrative examples of the four rule families considered in this work. Given the same input vector, the additive rule computes a weighted sum, the conjunctive and disjunctive rules return the minimum and maximum threshold margins ($x_i-\tau_i$), respectively, and the lexicographic rule returns the margin of the first threshold-satisfying input in priority order.}
    \label{fig:rule_families}
\end{figure}

\paragraph{Rule families.}
The four rule families correspond to four common attribute-combination
strategies:
\begin{itemize}
    \item \textbf{Additive.}
    Relevant fields are combined through a weighted sum. A high value on one
    field can compensate for a low value on another \citep{payne1993adaptive}.

    \item \textbf{Conjunctive.}
    All relevant fields must satisfy their conditions. A single field that
    falls below its threshold can limit the overall outcome
    \citep{einhorn1971noncompensatory}.

    \item \textbf{Disjunctive.}
    Satisfying any one relevant condition can be sufficient. Thus, one strong
    field may determine the outcome even when other fields are weak
    \citep{einhorn1970nonlinear}.

    \item \textbf{Lexicographic.}
    Relevant fields are checked in a fixed priority order. The
    highest-priority field is considered first, and lower-priority fields
    matter only when earlier fields are not decisive
    \citep{payne1993adaptive,gigerenzer1996reasoning}.
\end{itemize}
Figure~\ref{fig:rule_families} illustrates the structural difference among the
four families: additive rules \textit{aggregate} evidence, conjunctive rules
require \textit{all} conditions, disjunctive rules require \textit{any}
condition, and lexicographic rules follow a \textit{priority order}.

\paragraph{Case space.}
Each case is represented by a \(d\)-dimensional vector over predefined field
ranges:
\[
\mathbf{x}=(x_1,\ldots,x_d)\in\mathcal{X},
\qquad
\mathcal{X}=\prod_{j=1}^{d}[l_j,u_j],
\]
where only an unknown subset of fields
\(\mathcal{I}\subseteq\{1,\ldots,d\}\) is relevant to the decision.

\paragraph{Rule Function.}
The Rule Function first normalizes each relevant field to \([-1,1]\):
\[
\tilde{x}_j
=
2\frac{x_j-l_j}{u_j-l_j}-1.
\]
Each rule then combines the relevant fields into a scalar \textit{raw
evidence} value \(e\) according to one of four structural families.

For an \textit{additive} rule, evidence is aggregated across relevant fields:
\[
e
=
b+\sum_{j\in\mathcal{I}} w_j\tilde{x}_j,
\]
where \(w_j\) is the contribution of field \(j\) and \(b\) is a bias term.

For the remaining families, each relevant field has a threshold \(t_j\),
normalized analogously as \(\tilde{t}_j\), and a direction
\(d_j\in\{-1,+1\}\). Its oriented margin is
\[
m_j
=
d_j\bigl(\tilde{x}_j-\tilde{t}_j\bigr).
\]
Conjunctive and disjunctive rules aggregate these margins by their weakest and
strongest conditions, respectively:
\[
e_{\mathrm{conj}}
=
\min_{j\in\mathcal{I}} m_j,
\qquad
e_{\mathrm{disj}}
=
\max_{j\in\mathcal{I}} m_j.
\]

A \textit{lexicographic} rule additionally specifies a priority ordering $\pi$ over the relevant fields and an indifference band $\delta$.
The decisive field is determined by
\[
k^* =
\begin{cases}
\min\{k:|m_{\pi_k}|>\delta\},
& \text{if such }k\text{ exists},\\
|\mathcal{I}|, & \text{otherwise}.
\end{cases}
\]
The raw evidence is then
\[
e_{\mathrm{lex}}=m_{\pi_{k^*}}.
\]
Thus, the first decisive field determines the outcome, with the lowest-priority field used as a fallback.

Finally, a monotonic calibration function maps raw evidence to the intervention score:
\[
s=R_\theta(\mathbf{x})=C_\theta(e),
\qquad
C_\theta:\mathbb{R}\rightarrow[-1,1],
\]
where the calibration is determined from the pre-change evidence distribution, as detailed below.

\paragraph{Calibration from Evidence to Intervention Score.}
To obtain balanced action regions across heterogeneous rule families, we calibrate raw evidence using its distribution under the pre-change population.
Let $q_{25}$, $q_{50}$, and $q_{75}$ denote the empirical quartiles of raw evidence sampled from this population.
We map these quartiles to the three action boundaries $-0.5$, $0$, and $0.5$ using a monotonic piecewise-linear function:
\[
C_\theta(e) =
\operatorname{clip}_{[-1,1]}
\begin{cases}
-\frac{1}{2}+\frac{e-q_{25}}{2(q_{50}-q_{25})},
& e\leq q_{50},\\[4pt]
\frac{e-q_{50}}{2(q_{75}-q_{50})},
& e>q_{50}.
\end{cases}
\]
We require $q_{25}<q_{50}<q_{75}$ to ensure a well-defined mapping.
This calibration approximately balances the four optimal actions under the pre-change population.
For dynamic episodes, the calibration is fitted once and held fixed after environmental changes, preserving a consistent evidence-to-score mapping throughout the episode.

\paragraph{Utility Function.}
The Utility Function maps the intervention score to the utilities of the four
actions:
\[
\mathcal{U}(s)
=
\bigl(u_0(s),u_1(s),u_2(s),u_3(s)\bigr)
\in\mathbb{R}^4,
\]
where \(u_k(s)\) is the utility obtained by choosing action \(a_k\). Each
action is associated with a fixed score anchor
\(\mathbf{c}=(-0.75,-0.25,0.25,0.75)\), corresponding to increasing
intervention intensity. For each action \(a_k\), \(k=0,\ldots,3\), we define
\[
u_k(s)
=
\frac{1}{3}\left(
4\,
\frac{\exp[-\beta(s-c_k)^2]}
{\sum_{j=0}^{3}\exp[-\beta(s-c_j)^2]}
-1
\right),
\]
where \(\beta=8\) is fixed across all tasks. This construction gives higher
utility to actions whose anchors are closer to the intervention score. The
preferred action changes at the midpoints between adjacent anchors,
\(s\in\{-0.5,0,0.5\}\), yielding four equal-sized decision regions, and the
utilities sum to zero for every \(s\).

\section{Data Generation Details}
\label{app:data}

\subsection{Semantic Field Catalog Construction}

\paragraph{Catalog Specifications.}
The semantic field catalog is constructed using a predefined schema and task blueprints. Each blueprint specifies the task scenario, agent role, and four ordered actions, while the schema defines the required field attributes, rule-family specifications, cross-field constraints, and admissible environmental changes. LLM-assisted synthesis extends these predefined specifications with task-specific field semantics and numerical parameters.

\paragraph{Component Synthesis.}
For each task blueprint, we construct the semantic field catalog through the following component-level generation steps:

\begin{itemize}
    \item \textbf{Field Discovery.} The LLM proposes a set of candidate observable fields relevant to the task scenario, including field names, descriptions, units, and basic directional information.

    \item \textbf{Field Semantics.} For each retained field, the LLM specifies its applicability to different rule families, directional effects on intervention decisions, and admissible direction reversals with corresponding semantic events.

    \item \textbf{Field Parameters.} The LLM assigns numerical ranges, sampling distributions, and associated parameters to the accepted fields, establishing the field population used for subsequent episode generation.

    \item \textbf{Lexicographic Orders.} Based on the established field set, the LLM generates admissible priority orders for lexicographic decision rules.

    \item \textbf{Cross-Field Constraints.} The LLM defines feasibility constraints among fields to exclude invalid or contradictory combinations of field values.

    \item \textbf{Cross-Field Dynamics.} The LLM specifies admissible field replacement relations, including replacement groups, directional transitions, and coupling relationships for dynamic episodes.
\end{itemize}

\noindent Detailed generation prompts and output specifications are provided in Appendix~\ref{app:data_prompts}.

\paragraph{Validation and LLM Audit.}
Each generated component undergoes programmatic validation against the predefined schema and component-specific constraints, followed by LLM-based semantic auditing. Components failing these checks are revised and revalidated or rejected. Cross-component consistency is further verified to ensure valid field references and compatible semantic specifications. Only accepted components are integrated into the finalized semantic field catalog for subsequent episode generation.

\subsection{Episode Generation}

Episode generation consists of two steps: searching for valid field and rule configurations, followed by sampling evaluation cases and organizing them into batches.

\subsubsection{Field Configuration Search}

\paragraph{Candidate Enumeration.}
For each task and rule family, we enumerate candidate combinations of rule-relevant fields from the semantic field catalog. The enumeration follows a seed-determined bounded traversal of the admissible combinations. For dynamic episodes, candidates additionally incorporate admissible field replacements and direction reversals. Configurations violating cross-field feasibility or rule applicability constraints are discarded before numerical evaluation.

\paragraph{Numerical Certification.}
For each structurally valid configuration, we instantiate the corresponding latent dynamics and evaluate their properties over the feasible case population. Static candidates are calibrated and checked for non-degenerate optimal-action distributions. Dynamic candidates undergo three stages of numerical certification: initial screening, refinement, and held-out verification, using 256, 2048, and 16384 sampled cases, respectively. We evaluate the pre- and post-transition dynamics under a shared calibration, requiring adequate action coverage, moderate changes in optimal-action assignments, and feasible target action transitions. Specifically, the natural rate of changed optimal actions is constrained to 0.25--0.75, with marginal optimal-action proportions between 0.15 and 0.35.

\paragraph{Semantic Review.}
After numerical certification, additional distractor fields are selected to complete the observable field configuration. An LLM-based reviewer then assesses the semantic consistency of the selected fields and environmental dynamics with the task scenario. Candidates with invalid field configurations are rejected. Algorithm~\ref{alg:episode_generation} summarizes the construction procedure, and the review prompt specification are provided in Appendix~\ref{app:data_prompts}.

\begin{algorithm*}[t]
\caption{Field Configuration Enumeration and Validation}
\label{alg:episode_generation}
\begin{algorithmic}[1]
\Require Semantic field catalog \(\mathcal{F}\), constraints \(\Gamma,\Lambda\), generation cells \(\mathcal{G}\), quotas \(q\)
\Ensure Accepted configurations \(\mathcal{C}\)
\State \(\mathcal{C}\gets\varnothing\)
\ForAll{eligible generation cells \(g\in\mathcal{G}\)}
    \ForAll{candidate cores \(c\in\textsc{EnumerateCores}(g)\)}
        \If{\(\neg\textsc{StructurallyFeasible}(c,\Gamma,\Lambda)\)}
            \State \textbf{continue}
        \EndIf
        \State \((\mathcal{R}_{\mathcal{E}},\Delta_{\mathcal{E}})\gets\textsc{InstantiateDynamics}(c,g)\)
        \If{\(\neg\textsc{NumericallyValid}(c,\mathcal{R}_{\mathcal{E}},\Delta_{\mathcal{E}})\)}
            \State \textbf{continue}
        \EndIf
        \State \(\mathcal{F}_{\mathcal{E}}\gets\textsc{AddDistractors}(c,\mathcal{F})\)
        \State \(r\gets\textsc{SemanticReview}(\mathcal{F}_{\mathcal{E}},\mathcal{R}_{\mathcal{E}},\Delta_{\mathcal{E}})\)
        \While{\(r=\texttt{reject\_distractors}\) and retries remain}
            \State \(\mathcal{F}_{\mathcal{E}}\gets\textsc{ReselectDistractors}(c,\mathcal{F})\)
            \State \(r\gets\textsc{SemanticReview}(\mathcal{F}_{\mathcal{E}},\mathcal{R}_{\mathcal{E}},\Delta_{\mathcal{E}})\)
        \EndWhile
        \If{\(r=\texttt{accept}\)}
            \State Add \((\mathcal{F}_{\mathcal{E}},\mathcal{R}_{\mathcal{E}},\Delta_{\mathcal{E}})\) to \(\mathcal{C}\)
            \State Update remaining quotas
        \EndIf
        \If{\(\textsc{QuotaSatisfied}(g,q)\)}
            \State \textbf{break}
        \EndIf
    \EndFor
\EndFor
\State \Return \(\mathcal{C}\)
\end{algorithmic}
\end{algorithm*}

\subsubsection{Evaluation Case Sampling}

For each accepted configuration, evaluation cases \(\mathbf{x}\) are sampled from the specified field distributions. The applicable latent dynamics determine the action utilities \(\mathcal{U}(R_\theta(\mathbf{x}))\), and the resulting cases are organized into evaluation batches.

\paragraph{Static Setting.}
For static episodes, we independently sample cases from the field distributions conditioned on the cross-field feasibility constraints, and randomly organize them into evaluation batches. All cases are evaluated under the same latent dynamics.

\paragraph{Dynamic Setting.}
For dynamic episodes, we first sample a shared pre-transition prefix under the initial dynamics, followed by multiple post-transition branches. The post-transition sampling includes an IID control branch drawn from the feasible case population and several targeted branches with specified optimal-action change rates. For each targeted branch, cases are sampled from pre/post action-transition groups identified during numerical certification to achieve the desired change rate. The resulting branches share the same pre-transition cases but differ in their post-transition case distributions.

Algorithm~\ref{alg:case_sampling} summarizes the evaluation case sampling and batch assembly procedure.

\begin{algorithm*}[t]
\caption{Evaluation Case Sampling and Batch Assembly}
\label{alg:case_sampling}
\begin{algorithmic}[1]
\Require Accepted configuration \((\mathcal{F}_{\mathcal{E}},\mathcal{R}_{\mathcal{E}},\Delta_{\mathcal{E}})\), constraints \(\Gamma\), mode \(m\), number of batches \(T\), batch size \(b\), target rates \(\mathcal{Q}\)
\Ensure Evaluation batches \(\mathcal{B}_{\mathcal{E}}\)
\If{\(m=\texttt{static}\)}
    \State \(X\gets\textsc{SampleFeasible}(\mathcal{F}_{\mathcal{E}},\Gamma,Tb)\)
    \State \(\mathcal{B}_{\mathcal{E}}\gets\textsc{BatchWithUtilities}(X,R_\theta,b)\)
\Else
    \State Select an admissible transition boundary \(\tau\)
    \State \(X_{\mathrm{pre}}\gets\textsc{SampleFeasible}(\mathcal{F}_{\mathcal{E}},\Gamma,\tau b)\)
    \State \(B_{\mathrm{pre}}\gets\textsc{BatchWithUtilities}(X_{\mathrm{pre}},R_{\theta_{\mathrm{pre}}},b)\)
    \ForAll{post-transition branches \(r\in\{\texttt{IID}\}\cup\mathcal{Q}\)}
        \If{\(r=\texttt{IID}\)}
            \State \(X_{\mathrm{post}}^r\gets\textsc{SampleFeasible}(\mathcal{F}_{\mathcal{E}},\Gamma,(T-\tau)b)\)
        \Else
            \State \(X_{\mathrm{post}}^r\gets\textsc{SampleTransitionBuckets}(r,(T-\tau)b)\)
        \EndIf
        \State \(R^r\gets\textsc{ApplicableRule}(r,\mathcal{R}_{\mathcal{E}},\Delta_{\mathcal{E}})\)
        \State \(B_{\mathrm{post}}^r\gets\textsc{BatchWithUtilities}(X_{\mathrm{post}}^r,R^r,b)\)
    \EndFor
    \State \(\mathcal{B}_{\mathcal{E}}\gets(B_{\mathrm{pre}},\{B_{\mathrm{post}}^r\}_r)\)
\EndIf
\State \Return \(\mathcal{B}_{\mathcal{E}}\)
\end{algorithmic}
\end{algorithm*}

\subsection{Dataset-Level Review and Selection}
\label{app:data_review}

After episode generation, a separate LLM-based quality review evaluates each candidate along three dimensions: field realism, plausibility of dynamic rule transitions, and distractor quality. Episodes must exceed the minimum score threshold on every dimension to qualify for selection. Qualified episodes are ranked by their aggregate review scores within each domain--rule-family combination, and equal numbers are selected from these groups to maintain dataset balance.

The selected episodes form the final evaluation dataset. Detailed review criteria and prompts are provided in Appendix~\ref{app:data_prompts}.

\subsection{Dataset Configuration}
\label{app:dataset_configuration}

\paragraph{Dataset Composition.}
Our evaluation dataset consists exclusively of dynamic episodes, balanced across all domain--rule-family combinations. We generate 384 candidate episodes, with 24 per combination, and retain the 256 highest-scoring qualified episodes through the dataset-level review and selection procedure, yielding 16 episodes per combination.

\paragraph{Evaluation Batches.}
Each episode contains 12 action batches, with 8 evaluation cases per batch. The \dynamic{} occurs after batch 5, 6, or 7, with the preceding batches shared across all post-transition branches. We construct one IID control branch and three targeted branches with optimal-action change rates of 0.25, 0.50, and 0.75. Each complete trajectory contains 96 evaluation cases. The evaluation protocol additionally specifies a probe budget of 16 per batch, a maximum of 32 probe cases, and a total probe budget of 32.

\paragraph{Training Set.}
We additionally provide a training set containing 1,000 episodes, generated using the same data generation and validation pipeline as the evaluation dataset.

\subsection{LLM Prompt Specifications}
\label{app:data_prompts}

\paragraph{Semantic Field Catalog Construction.}
Figures~\ref{fig:prompt-field-discovery}, \ref{fig:prompt-field-semantics}, and \ref{fig:prompt-field-parameters} present the system prompts for field discovery, semantic specification, and numerical parameter calibration, respectively. Figures~\ref{fig:prompt-lexicographic-orders}, \ref{fig:prompt-field-constraints}, and \ref{fig:prompt-field-replacements} provide the prompts for constructing lexicographic priority orders, cross-field feasibility constraints, and admissible field replacements.

\paragraph{Episode and Dataset Review.}
Figure~\ref{fig:prompt-candidate-semantic-review} presents the prompt used to assess candidate episode configurations and distractor fields. Figure~\ref{fig:prompt-final-dataset-review} presents the final dataset-level review prompt, which scores episodes on field realism, rule-change realism, and distractor quality.

\section{Human Study}
\label{app:human-study}

To assess the semantic plausibility of our LLM-assisted generated tasks, we randomly sampled 40 episodes from the 256 dynamic episodes and invited three human annotators to evaluate each case along three dimensions: 
(i) scenario plausibility, assessing whether the task setting and available actions are realistic; 
(ii) factor replacement, assessing whether the contextual changes in relevant fields and their underlying reasons justify updating existing information; and 
(iii) effect reversal, assessing whether the changes reasonably justify reversing the influence of an attribute on decision-making. 
Each dimension was rated on a five-point Likert scale, with scores of 4 or 5 indicating positive assessments. As shown in Table~\ref{tab:human_study}, all three dimensions received favorable evaluations, with mean scores above 4.0. Notably, at least 90\% of the evaluated cases received positive ratings from a majority of annotators across all dimensions, supporting the overall semantic plausibility of our generated tasks and dynamic mechanisms.

\begin{table}[t]
    \centering
    \footnotesize
    \setlength{\tabcolsep}{3pt}
    \resizebox{\columnwidth}{!}{
        \begin{tabular}{@{}lccc@{}}
            \toprule
            \textbf{Dimension} &
            \textbf{Mean $\pm$ SD} &
            \textbf{Pos. (\%)} &
            \textbf{Maj. (\%)} \\
            \midrule
            Scenario Plausibility   & 4.95 $\pm$ 0.22 & 100.0 & 100.0 \\
            Information Replacement & 4.56 $\pm$ 0.61 & 94.2 & 100.0 \\
            Effect Reversal         & 4.09 $\pm$ 0.73 & 77.5 & 90.0 \\
            \bottomrule
        \end{tabular}
    }
    \caption{Human evaluation of generated task plausibility. Positive denotes the percentage of ratings $\geq 4$, while Majority denotes the percentage of cases rated $\geq 4$ by at least two annotators.}
    \vspace{-0.5em}
    \label{tab:human_study}
\end{table}

\section{Detailed Runtime Settings}
\label{app:runtime_settings}

\paragraph{Runtime configs.}
Table~\ref{tab:runtime-settings} lists the runtime configuration shared by every model and episode.
Each episode presents its evaluation cases as a fixed sequence of batches, and \dynamic{} is scheduled after a batch drawn approximately uniformly from the middle of that sequence.
Within each batch window the agent may probe at most once, each probe call is capped in size, and probing is always optional because the upcoming batch can also be revealed directly.
Action submissions may abstain on individual cases, and unanswered cases earn zero utility.
A rejected call may be repaired once per window before the environment applies the fallbacks described above; provider failures are retried with capped exponential backoff, and an over-length response is regenerated once before its window is skipped.
These limits are identical for all models, and no model profile overrides them.

\begin{table}[t]
  \centering
  \small
  \resizebox{\columnwidth}{!}{
  \begin{tabular}{ll}
    \toprule
    Setting & Value \\
    \midrule
    \multicolumn{2}{l}{\textit{Episode layout}} \\
    Action batches per episode & 12 \\
    Cases per action batch & 8 \\
    Candidate actions per case & 4 \\
    Input fields per case & 8 \\
    \dynamic{} after batch & 5, 6, or 7 \\
    \midrule
    \multicolumn{2}{l}{\textit{Probing and answering}} \\
    Max probe cases per \texttt{probe\_batch} call & 16 \\
    Probe calls per batch window & $\leq 1$ \\
    Min probe cases per \texttt{probe\_batch} call & 1 \\
    Probe-free reveal (\texttt{request\_next\_batch}) & allowed \\
    Per-case abstention & allowed \\
    \midrule
    \multicolumn{2}{l}{\textit{Failure handling}} \\
    Repair attempts per rejected call & 1 \\
    Provider retries per request & 5 (backoff 2--16\,s) \\
    Regenerations after a truncated response & 1 \\
    \bottomrule
  \end{tabular}
  }
  \caption{Interaction and failure-handling configuration of the benchmark
  runtime. Every value is fixed across episodes, models, and settings. The
  fallbacks that apply once a limit is exhausted are described in the text.}
  \vspace{-0.5em}
  \label{tab:runtime-settings}
\end{table}

\paragraph{Runtime Failure Handling.}
We distinguish protocol errors from provider failures.
\textbf{(1) Repair.} Invalid agent responses, including malformed JSON, schema violations, and invalid tool calls, trigger an error message in the agent's context and one additional interaction for correction, without advancing the environment.
If the repair fails, the current exploration window is skipped, or unanswered evaluation cases receive zero utility and are counted as incorrect.
Mandatory probing requirements are waived for the affected window.
\textbf{(2) Retry.} Provider and network failures are retried with capped exponential backoff, without affecting the agent's interaction history or repair budget.
Exhausted retries mark the run as failed.
Truncated responses receive one additional generation attempt, after which repeated truncation triggers the corresponding fallback.

\paragraph{Runtime Prompts.}
Figure~\ref{fig:inference-prompt} shows the initial prompt that opens every
episode under Default, and Figures~\ref{fig:prompt-probe-batch}, \ref{fig:prompt-request-next-batch},
\ref{fig:prompt-submit-action-batch}, and \ref{fig:prompt-calculator} show the
environment feedback that answers each of the four interaction tools. Angle
brackets mark content that varies across episodes, such as the role
assignment, the action names, and the field descriptions, while ellipses and
bracketed elision notes stand for entries that repeat a displayed structure.
When an episode defines feasibility constraints, they appear as a constraint
list after the field descriptions in the first user message. Every environment
message ends with the latest \texttt{Current state} card, which is the
authoritative snapshot of tool usability, cumulative utility, pending cases,
and remaining action batches. A Request Next Batch call is answered through
this card alone, because revealing a batch without probing produces no
separate result payload. After each submitted action batch, the environment
rewrites the retained history as described in the system message, so later
turns observe the scoreboard form of earlier outcomes instead of the original
messages. Feedback for rejected calls is omitted from these figures, since a
rejection only triggers a repair opportunity and leaves the episode state
unchanged.

\section{Model Configs}
\label{app:model_setup}

\begin{table}[t]
  \centering
  \small
  \resizebox{\columnwidth}{!}{%
  \begin{tabular}{lccc}
    \toprule
    Model Name & Effort & Thinking Budget & Max Output Tokens \\
    \midrule
    \textit{Llama-3.1-8B} & default & -- & 8192 \\
    \textit{Qwen3.5-9B} & default & 4096 & 8192 \\
    \textit{Qwen3.8-27B} & medium & -- & 8192 \\
    \textit{Gemma-4-12B} & high & 4096 & 8192 \\
    \textit{DeepSeek-V4.1-Flash} & high & -- & 32768 \\
    \textit{GPT-6-Luna} & high & -- & 32768 \\
    \textit{GPT-6-Sol} & high & -- & 32768 \\
    \textit{Gemini-3.8-Flash} & high & -- & 32768 \\
    \bottomrule
  \end{tabular}}
  \caption{Configuration snapshot of every reported run, frozen into the
  run directory at launch. A dash means the setting is unset.}
  \vspace{-0.5em}
  \label{tab:model-configs}
\end{table}

\paragraph{Reasoning Configs.} Table~\ref{tab:model-configs} lists the configuration snapshot that the benchmark runner froze into each reported run at launch.
All eight models must answer in JSON, and temperature is 0.0 wherever the endpoint acceptsit. 
The \emph{default} effort of \textit{Llama-3.1-8B} and \textit{Qwen3.5-9B} is not a lowered reasoning setting, since the former has no reasoning mode and the latter always reasons.

\paragraph{Resource Used.} We use 4x Nvidia L20 GPU to host the open-source small models. Large models and close-source models are accessed through their official APIs.

\section{Formal Definitions of Evaluation Metrics}
\label{app:metric_definitions}

\paragraph{Calibrated Accuracy.}
For evaluation case $i$, let $u_i(a)$ denote the utility of action $a$ and
$\hat a_i$ the action selected by the agent. We first compute
\[
p_i =
\frac{
u_i(\hat a_i)-\min_{a\in\mathcal A}u_i(a)
}{
\max_{a\in\mathcal A}u_i(a)-\min_{a\in\mathcal A}u_i(a)
}.
\]
The corresponding uniformly random baseline is
\[
b_i =
\frac{1}{|\mathcal A|}
\sum_{a\in\mathcal A}
\frac{
u_i(a)-\min_{a'}u_i(a')
}{
\max_{a'}u_i(a')-\min_{a'}u_i(a')
}.
\]
Let $\bar p$ and $\bar b$ denote their averages over evaluation cases.
We report
\[
\mathrm{CA}
=
\frac{\bar p-\bar b}{1-\bar b},
\]
such that a uniformly random policy obtains $0$ in expectation and the
oracle policy obtains $1$.
Unlike binary accuracy, CA preserves information about the relative quality
of suboptimal decisions~\citep{liu2026naacl}.

\paragraph{Mean Unique Probes.}
Every probe instantiates a complete context vector on the agent-facing value
grid (integer fields are exact; decimal fields are resolved at two decimal
places), so two probes are duplicates if and only if their vectors coincide
exactly. Splitting an episode at the rule change, let
$\mathcal S^{\mathrm{pre}}$ and $\mathcal S^{\mathrm{post}}$ be the sets of
\emph{distinct} probe vectors queried in the pre- and post-change phases;
repeated queries of the same point within a phase are counted once~\citep{liu-etal-2024-gprooft}.
A post-change probe that revisits a pre-change point is \emph{not}
deduplicated: under the new rule it returns fresh evidence, so it still
counts as unique (this overlap is measured separately below). On branches
without a rule change, the episode is treated as a single phase. We report
\[
\frac{1}{N}\sum_{i=1}^{N}
\Big(
\big|\mathcal S^{\mathrm{pre}}_i\big|
+
\big|\mathcal S^{\mathrm{post}}_i\big|
\Big).
\]

\paragraph{Factor Recognition.}
The auxiliary assessments described in
Section~\ref{sec:runtime_protocol} allow us to isolate rule induction from
action execution.
Each assessment reports a predicted relevant-field set
$\hat{\mathcal I}$; reported directions are not scored here and enter only
through Change Recognition below.
For each phase $t \in \{\mathrm{pre}, \mathrm{post}\}$, we pool counts over
all episodes with a successfully completed assessment,
$\mathrm{TP}_t = \sum_i |\hat{\mathcal I}_{i,t} \cap \mathcal I_{i,t}|$,
$\mathrm{FP}_t = \sum_i |\hat{\mathcal I}_{i,t} \setminus \mathcal I_{i,t}|$,
$\mathrm{FN}_t = \sum_i |\mathcal I_{i,t} \setminus \hat{\mathcal I}_{i,t}|$,
and compute a single micro-averaged
\[
\mathrm{F1}_t
=
\frac{2\,\mathrm{TP}_t}
{2\,\mathrm{TP}_t + \mathrm{FP}_t + \mathrm{FN}_t}.
\]
Because every episode has exactly three relevant fields, micro and macro
recall coincide, and pooling chiefly smooths the coarse per-episode
granularity of F1.
The Factor Recognition score is
$\tfrac{1}{2}\big(\mathrm{F1}_{\mathrm{pre}} + \mathrm{F1}_{\mathrm{post}}\big)$.
Failed or missing assessments never enter a pool; on branches without a rule
change we pool the single final assessment instead.

\paragraph{Change Recognition.}
We represent the rule transition as a set of atomic changes.
A field replacement contributes a removal atom
$(f^{\mathrm{out}}, d^{\mathrm{pre}}_{f^{\mathrm{out}}})$ and an addition atom
$(f^{\mathrm{in}}, d^{\mathrm{post}}_{f^{\mathrm{in}}})$, each pairing the
field with its direction in the respective rule, while a direction reversal
contributes a reversal atom
$(f, d^{\mathrm{pre}}_f, d^{\mathrm{post}}_f)$.
Every transition in our benchmark is compound, so the ground-truth change set
$\mathcal C^{*}$ always contains exactly three atoms.
We construct the predicted change set $\hat{\mathcal C}$ from the model's two
assessments by the same construction---fields leaving the reported set yield
removal atoms, fields entering it yield addition atoms, and retained fields
whose reported direction flips yield reversal atoms---and score it against
$\mathcal C^{*}$ by set F1 with strict matching: a reversal atom is correct
only if the field and \emph{both} its pre- and post-change directions match.
The Change Recognition score is the mean per-episode F1 over episodes in
which both assessments completed successfully, and we report the
paired-assessment completion rate alongside it.

\paragraph{Mean Re-probes after Change.}
Using the probe-point sets defined above, we report
\[
\frac{1}{N}\sum_{i=1}^{N}
\big|\mathcal S^{\mathrm{post}}_i \cap \mathcal S^{\mathrm{pre}}_i\big|,
\]
the number of \emph{distinct} pre-change probe points that the agent queries
again after the rule change; repeated visits to the same point count once.
It is the exact complement of Mean Unique Probes on post-change probes:
\[
|\mathcal S^{\mathrm{post}}|
= |\mathcal S^{\mathrm{post}} \setminus \mathcal S^{\mathrm{pre}}|
+ |\mathcal S^{\mathrm{post}} \cap \mathcal S^{\mathrm{pre}}|,
\]
splitting post-change exploration into novel points and revisited points.
The metric is defined only on branches with a rule change and is bounded
above by $|\mathcal S^{\mathrm{pre}}|$.

\section{Analysis Results}
\label{app:analysis_results}

\subsection{Correlation Analysis}
\label{app:correlation_analysis}

Table~\ref{tab:metric-correlation} quantifies how strongly exploration
volume and the two recognition metrics co-vary with the end-to-end
outcomes. Each cell correlates the eight model means from
Table~\ref{tab:main_results}, so the table measures whether a metric
separates strong models from weak ones, and the Avg.\ columns average the
EA and CA correlations.

\begin{table}[t]
  \centering
  \resizebox{\columnwidth}{!}{%
  \begin{tabular}{lcccccc}
    \toprule
    & \multicolumn{3}{c}{Pearson $r$} & \multicolumn{3}{c}{Spearman $\rho$} \\
    \cmidrule(lr){2-4}\cmidrule(lr){5-7}
    Metric & EA & CA & Avg. & EA & CA & Avg. \\
    \midrule
    Mean Uni. Probes & 0.912 & 0.916 & 0.914 & 0.881 & 0.905 & 0.893 \\
    Factor Rec. & 0.913 & 0.922 & 0.917 & \textbf{0.929} & \textbf{0.952} & \textbf{0.940} \\
    Change Rec. & \textbf{0.937} & \textbf{0.928} & \textbf{0.932} & 0.833 & 0.905 & 0.869 \\
    \bottomrule
  \end{tabular}}
  \caption{Pearson $r$ and Spearman $\rho$ of Mean Unique Probes and the
  two recognition metrics with the end-to-end outcomes, computed across
  the eight model means. Bold marks the strongest correlation in each
  column.}
  \label{tab:metric-correlation}
\end{table}

\subsection{Full Results for Analysis}
\label{app:full_analysis}

\begin{table}[t]
  \centering
  \small
  \resizebox{\columnwidth}{!}{%
  \begin{tabular}{lcccc}
    \toprule
    & \multicolumn{2}{c}{Stationary}
    & \multicolumn{2}{c}{Stationary-IID} \\
    \cmidrule(lr){2-3}\cmidrule(lr){4-5}
    Model & EA & CA & EA & CA \\
    \midrule
    DeepSeek-V4.1-Flash & 69.19 & 67.60 & 68.85 & 67.39 \\
    Qwen3.8-27B         & 54.74 & 50.57 & 55.60 & 51.37 \\
    GPT-6-Luna          & 50.26 & 41.14 & 50.82 & 42.23 \\
    Gemma-4-12B         & 39.37 & 22.78 & 40.32 & 24.86 \\
    Qwen3.5-9B          & 32.15 & 12.41 & 33.29 & 13.86 \\
    Llama-3.1-8B        & 30.90 &  8.74 & 31.53 &  9.74 \\
    \midrule
    Mean                & 46.10 & 33.87 & 46.73 & 34.91 \\
    \bottomrule
  \end{tabular}}
  \caption{Performance results comparing \textit{stationary} and  \textit{stationary setting with IID sampling}.}
  \vspace{-0.5em}
  \label{tab:ablation-sampling}
\end{table}
\begin{table}[t]
  \centering
  \resizebox{\columnwidth}{!}{%
  \begin{tabular}{lcc}
    \toprule
    Model & Factor Replacement & Direction Reversal \\
    \midrule
    \textit{Llama-3.1-8B} & \textbf{6.9} & 5.9 \\
    \textit{Qwen3.5-9B} & \textbf{21.1} & 2.4 \\
    \textit{Qwen3.8-27B} & \textbf{31.7} & 6.0 \\
    \textit{Gemma-4-12B} & \textbf{14.7} & 2.0 \\
    \textit{DeepSeek-V4.1-Flash} & \textbf{49.2} & 19.6 \\
    \textit{GPT-6-Luna} & \textbf{13.1} & 1.6 \\
    \textit{GPT-6-Sol} & \textbf{57.6} & 21.5 \\
    \textit{Gemini-3.8-Flash} & \textbf{73.6} & 27.7 \\
    \midrule
    Pooled & \textbf{33.5} & 10.8 \\
    \bottomrule
  \end{tabular}}
  \caption{Change Recognition decomposed by atom type on \textit{default}. The larger of the two rates is in bold. The replacement rate exceeds the reversal rate at $p<10^{-70}$ under a paired Wilcoxon
  signed-rank test.}
  \label{tab:atom-breakdown}
\end{table}
\begin{table}[t]
  \centering
  \small
  \resizebox{\columnwidth}{!}{%
  \begin{tabular}{lccc}
    \toprule
    & \multicolumn{2}{c}{End-to-end Outcome} & Exploration Behavior \\
    \cmidrule(lr){2-3}\cmidrule(lr){4-4}
    Model Name & EA (\%) $\uparrow$ & CA (\%) $\uparrow$ & Mean Uni.\ Probes \\
    \midrule
    \textit{Llama-3.1-8B} & 30.90 & 8.74 & 4.53 \\
    \textit{Qwen3.5-9B} & 32.15 & 12.41 & 6.76 \\
    \textit{Qwen3.8-27B} & 54.74 & 50.57 & 33.31 \\
    \textit{Gemma-4-12B} & 39.37 & 22.78 & 12.28 \\
    \textit{DeepSeek-V4.1-Flash} & \textbf{69.19} & \textbf{67.60} & 85.36 \\
    \textit{GPT-6-Luna} & 50.26 & 41.14 & 21.11 \\
    \textit{GPT-6-Sol} & \underline{65.60} & \underline{62.46} & 57.02 \\
    \bottomrule
  \end{tabular}}
  \caption{Evaluation results under the Stationary control, where the rule never changes. Mean Re-probes and Adaptation Performance are undefined in this setting. Gemini-3.8-Flash has no diagnostic runs and is omitted.}
  \label{tab:stationary-results}
\end{table}

Tables~\ref{tab:stationary-results}, \ref{tab:evidence-given-results}, and
\ref{tab:forced-probe-results} report the full per-model results under the
\emph{Stationary}, \emph{Evidence Given}, and \emph{Forced-Probe} settings
introduced in Section~\ref{sec:analysis}, complementing the \emph{default}
results in Table~\ref{tab:main_results}. Since \emph{Stationary} removes
the rule change, Mean Re-probes and the two Adaptation Performance metrics
are undefined, and the table reports only the End-to-end Outcome metrics
and Mean Unique Probes. \emph{Evidence Given} and \emph{Forced-Probe}
retain the rule change and report the same six metrics as the main table.
\textit{Gemini-3.8-Flash} has no diagnostic runs and is omitted from all three
tables.

\subsection{Significance Analysis}
\label{app:significance_analysis}

Table~\ref{tab:setting-significance} reports paired Wilcoxon signed-rank
tests that compare each diagnostic setting with \emph{default} on
per-episode Calibrated Accuracy. The gains under \emph{Stationary} and
\emph{Evidence Given} are significant at the model level, and the
\emph{Evidence Given} gain is positive and significant for nearly every
model. The \emph{Forced-Probe} effect is mixed across models and its
model-level mean is not significant.

\section{Ablation Studies}
\label{app:ablation}

\paragraph{Ablation on Evaluation-case Sampling.}
We further performs an ablation study on \textit{Stationary Setting with IID Sampling}.
It follows \textit{Stationary} setting except that the evaluation cases are sampled randomly rather than controlled for 50\% flip rate.
Table~\ref{tab:ablation-sampling} compares Stationary with a Stationary-IID variant whose evaluation cases are sampled randomly.
The two differ by at most 2.1 CA points per model and 1.0 point on average, so the sampling scheme has little effect and is not significant.
The difficulty of \textit{default}  setting thus stems from the unannounced \dynamic{} and self-collecting evidence, not from how evaluation cases are distributed.

\section{Change Recognition Decomposition Results}
\label{app:episode-error-analysis}

\paragraph{Assessment outcomes.}
Table~\ref{tab:factor-outcomes} decomposes Factor Recognition into
per-assessment outcomes on the dynamic branch of Default, reported
separately for the pre-change and post-change phases. Each completed
assessment falls into exactly one of four exhaustive categories, depending
on whether the reported factor set matches the ground truth exactly, omits
relevant factors without adding irrelevant ones, adds irrelevant factors
without omissions, or mixes both error types. Assessments reporting no
factors at all count as omission only.

\paragraph{Atom decomposition.}
Table~\ref{tab:atom-breakdown} decomposes Change Recognition on the
dynamic branch of Default by atom type and reports the share of
ground-truth atoms recovered under strict matching. Every transition
contributes two replacement atoms, one removal and one addition, and a
single reversal atom, so the two columns average over different numbers of
atoms per episode.

\paragraph{Outcome categories.}
Table~\ref{tab:episode-outcomes} complements the per-atom breakdown in
Table~\ref{tab:atom-breakdown} with an episode-level view of the same episodes.
Each episode falls into exactly one of four categories under strict matching, depending on whether the agent recovered both replacement atoms together with the reversal, the full replacement alone, the reversal alone, or neither.
An episode counts only when the agent completed both factor assessments.
Because recovering the replacement requires both of its atoms while the reversal contributes one, the replacement-only column is stricter than the per-atom replacement rate in Table~\ref{tab:atom-breakdown}.
The unchanged-report column counts episodes whose post-change assessment repeats the pre-change one exactly, which recovers no atom and therefore overlaps with the neither column.

\section{Case Study}
\label{app:case_study}

\subsection{Data Case Study}
\label{app:data_case_study}

Figure~\ref{fig:data-case-study} presents a production scheduling instance, including the complete task specification, a concrete input case, and its reference output.
The hidden rules are included alongside the agent-facing data so that the reader can inspect which fields determine the decision and how their effects change.

\subsection{Error Case Study}
\label{app:error_case_study}

Figure~\ref{fig:error-case-study} follows \textit{GPT-6 Sol} under Default on the same instance.
We reproduce the complete exploration and action exchange for the first post-change batch, together with the pre-change and final factor assessments.
The agent updates the relevant factor set correctly but retains the old direction for line throughput trend, even after collecting probes that vary this field while holding the other inputs fixed.

\begin{table*}[t]
  \centering
  \resizebox{\linewidth}{!}{%
  \begin{tabular}{lccccccccr}
    \toprule
    & \multicolumn{4}{c}{Pre-change (\%)} & \multicolumn{4}{c}{Post-change (\%)} & \\
    \cmidrule(lr){2-5}\cmidrule(lr){6-9}
    Model & Exact & Omit only & Over-report only & Mixed & Exact & Omit only & Over-report only & Mixed & $n$ (pre/post) \\
    \midrule
    \textit{Llama-3.1-8B} & 0.4 & 0.0 & 60.0 & 39.6 & 0.0 & 0.4 & 77.0 & 22.7 & 255/256 \\
    \textit{Qwen3.5-9B} & 0.0 & 32.4 & 12.3 & 55.3 & 0.0 & 29.7 & 10.5 & 59.8 & 253/256 \\
    \textit{Qwen3.8-27B} & 29.1 & 65.7 & 1.6 & 3.5 & 2.4 & 44.0 & 4.0 & 49.6 & 254/252 \\
    \textit{Gemma-4-12B} & 10.9 & 48.8 & 18.0 & 22.3 & 0.4 & 17.3 & 19.2 & 63.1 & 256/255 \\
    \textit{DeepSeek-V4.1-Flash} & 46.1 & 49.6 & 2.0 & 2.3 & 18.0 & 30.2 & 9.4 & 42.4 & 256/255 \\
    \textit{GPT-6-Luna} & 21.1 & 59.4 & 8.6 & 10.9 & 0.0 & 19.5 & 9.4 & 71.1 & 256/256 \\
    \textit{GPT-6-Sol} & 38.3 & 47.3 & 8.6 & 5.9 & 17.6 & 37.5 & 10.9 & 34.0 & 256/256 \\
    \textit{Gemini-3.8-Flash} & 43.8 & 54.7 & 0.4 & 1.2 & 20.3 & 61.3 & 2.3 & 16.0 & 256/256 \\
    \midrule
    Pooled & 23.8 & 44.8 & 13.9 & 17.6 & 7.3 & 30.0 & 17.9 & 44.8 & 2042/2042 \\
    \bottomrule
  \end{tabular}}
  \caption{Factor Recognition decomposed into per-assessment outcomes on
  the dynamic branch of Default, reported as shares of completed
  assessments before and after \dynamic{}. Each assessment falls into
  exactly one of the four columns. $n$ lists the completed assessments in
  each phase, and the pooled row aggregates assessments across models.}
  \label{tab:factor-outcomes}
\end{table*}

\begin{table*}[t]
  \centering
  \resizebox{\linewidth}{!}{%
  \begin{tabular}{lcccccc}
    \toprule
    Model & Both & Replacement only & Reversal only & Neither & Unchanged report & $n$ \\
    \midrule
    \textit{Llama-3.1-8B} & 0.0 & 2.7 & 5.9 & 91.4 & 33.3 & 255 \\
    \textit{Qwen3.5-9B} & 0.0 & 11.9 & 2.4 & 85.8 & 29.6 & 253 \\
    \textit{Qwen3.8-27B} & 2.4 & 22.3 & 3.6 & 71.7 & 54.2 & 251 \\
    \textit{Gemma-4-12B} & 0.0 & 4.3 & 2.0 & 93.7 & 44.3 & 255 \\
    \textit{DeepSeek-V4.1-Flash} & 14.5 & 19.2 & 5.1 & 61.2 & 29.4 & 255 \\
    \textit{GPT-6-Luna} & 0.0 & 5.1 & 1.6 & 93.4 & 65.6 & 256 \\
    \textit{GPT-6-Sol} & 17.6 & 30.1 & 3.9 & 48.4 & 21.9 & 256 \\
    \textit{Gemini-3.8-Flash} & 19.5 & 45.7 & 8.2 & 26.6 & 12.1 & 256 \\
    \midrule
    Pooled & 6.8 & 17.7 & 4.1 & 71.5 & 36.3 & 2037 \\
    \bottomrule
  \end{tabular}}
  \caption{Episode-level decomposition of Change Recognition outcomes on
  the dynamic branch of Default, reported as shares of eligible episodes.
  Each episode falls into exactly one of the first four columns, while
  unchanged report overlaps with the neither column. $n$ counts eligible
  episodes, and the pooled row aggregates episodes across models.}
  \label{tab:episode-outcomes}
\end{table*}
\begin{table*}[t]
  \centering
  \small
  \resizebox{\linewidth}{!}{%
  \begin{tabular}{lcccccc}
    \toprule
    & \multicolumn{2}{c}{End-to-end Outcome} & \multicolumn{2}{c}{Exploration Behavior} & \multicolumn{2}{c}{Adaptation Performance} \\
    \cmidrule(lr){2-3}\cmidrule(lr){4-5}\cmidrule(lr){6-7}
    Model Name & EA (\%) $\uparrow$ & CA (\%) $\uparrow$ & Mean Uni.\ Probes & Mean Re-probes & Factor Rec.\ (\%) $\uparrow$ & Change Rec.\ (\%) $\uparrow$ \\
    \midrule
    \textit{Llama-3.1-8B} & 34.13 & 13.78 & 4.87 & 0.41 & 52.55 & 5.32 \\
    \textit{Qwen3.5-9B} & 43.30 & 28.18 & 4.72 & 0.29 & 40.91 & 15.39 \\
    \textit{Qwen3.8-27B} & 35.73 & 33.78 & 17.36 & 0.08 & 62.65 & 13.85 \\
    \textit{Gemma-4-12B} & 48.10 & 36.52 & 7.00 & 0.00 & 50.34 & 12.10 \\
    \textit{DeepSeek-V4.1-Flash} & \textbf{71.51} & \textbf{70.34} & 119.28 & 6.38 & \textbf{79.92} & \underline{43.65} \\
    \textit{GPT-6-Luna} & 56.93 & 51.53 & 23.92 & 0.34 & 59.73 & 9.91 \\
    \textit{GPT-6-Sol} & \underline{68.51} & \underline{66.57} & 79.08 & 2.55 & \underline{76.41} & \textbf{44.40} \\
    \bottomrule
  \end{tabular}}
  \caption{Evaluation results under the Evidence Given setting.}
  \label{tab:evidence-given-results}
\end{table*}
\begin{table*}[t]
  \centering
  \small
  \resizebox{\linewidth}{!}{%
  \begin{tabular}{lcccccc}
    \toprule
    & \multicolumn{2}{c}{End-to-end Outcome} & \multicolumn{2}{c}{Exploration Behavior} & \multicolumn{2}{c}{Adaptation Performance} \\
    \cmidrule(lr){2-3}\cmidrule(lr){4-5}\cmidrule(lr){6-7}
    Model Name & EA (\%) $\uparrow$ & CA (\%) $\uparrow$ & Mean Uni.\ Probes & Mean Re-probes & Factor Rec.\ (\%) $\uparrow$ & Change Rec.\ (\%) $\uparrow$ \\
    \midrule
    \textit{Llama-3.1-8B} & 27.83 & 4.07 & 19.81 & 2.05 & 51.10 & 7.74 \\
    \textit{Qwen3.5-9B} & 33.59 & 14.32 & 17.46 & 4.41 & 38.66 & 12.76 \\
    \textit{Qwen3.8-27B} & 47.98 & 38.69 & 58.09 & 8.01 & 66.34 & 38.70 \\
    \textit{Gemma-4-12B} & 38.96 & 22.02 & 26.25 & 5.37 & 52.10 & 9.59 \\
    \textit{DeepSeek-V4.1-Flash} & \underline{57.98} & \underline{53.27} & 133.05 & 18.68 & \textbf{78.54} & \underline{48.91} \\
    \textit{GPT-6-Luna} & 44.77 & 32.27 & 68.04 & 2.31 & 59.60 & 14.01 \\
    \textit{GPT-6-Sol} & \textbf{59.89} & \textbf{53.99} & 120.44 & 5.02 & \underline{73.61} & \textbf{50.66} \\
    \bottomrule
  \end{tabular}}
  \caption{Evaluation results under the Forced-Probe setting.}
  \label{tab:forced-probe-results}
\end{table*}
\begin{table*}[t]
  \centering
  \small
  \resizebox{\linewidth}{!}{%
  \begin{tabular}{lcccccc}
    \toprule
    & \multicolumn{2}{c}{Stationary vs.\ Default} & \multicolumn{2}{c}{Evidence Given vs.\ Default} & \multicolumn{2}{c}{Forced-Probe vs.\ Default} \\
    \cmidrule(lr){2-3}\cmidrule(lr){4-5}\cmidrule(lr){6-7}
    Model Name & $\Delta$CA & $p$ & $\Delta$CA & $p$ & $\Delta$CA & $p$ \\
    \midrule
    \textit{Llama-3.1-8B} & $+0.62$ & $0.294$ & $+5.65$ & $<0.001^{***}$ & $-4.06$ & $<0.001^{***}$ \\
    \textit{Qwen3.5-9B} & $+0.10$ & $0.893$ & $+15.87$ & $<0.001^{***}$ & $+2.01$ & $0.066$ \\
    \textit{Qwen3.8-27B} & $+12.85$ & $<0.001^{***}$ & $-3.94$ & $0.007^{**}$ & $+0.96$ & $0.618$ \\
    \textit{Gemma-4-12B} & $+2.19$ & $0.003^{**}$ & $+15.93$ & $<0.001^{***}$ & $+1.43$ & $0.132$ \\
    \textit{DeepSeek-V4.1-Flash} & $+14.74$ & $<0.001^{***}$ & $+17.47$ & $<0.001^{***}$ & $+0.40$ & $0.711$ \\
    \textit{GPT-6-Luna} & $+6.05$ & $<0.001^{***}$ & $+16.43$ & $<0.001^{***}$ & $-2.82$ & $0.029^{*}$ \\
    \textit{GPT-6-Sol} & $+0.58$ & $0.889$ & $+4.69$ & $0.004^{**}$ & $-7.88$ & $<0.001^{***}$ \\
    \midrule
    Mean & $+5.30$ & $0.016^{*}$ & $+10.30$ & $0.031^{*}$ & $-1.42$ & $0.578$ \\
    \bottomrule
  \end{tabular}}
  \caption{Paired Wilcoxon signed-rank tests on per-episode Calibrated
  Accuracy between each diagnostic setting and Default, over 256 paired
  episodes per model. $\Delta$CA is the mean difference in percentage
  points, positive when the diagnostic setting scores higher. The Mean row
  averages the per-model deltas and tests them at the model level. All
  seven models with diagnostic runs are reported, while the main text
  figure shows six of them for space.
  $^{*}\,p<0.05$, $^{**}\,p<0.01$, $^{***}\,p<0.001$.}
  \label{tab:setting-significance}
\end{table*}


\begin{template*}[title={Data Case Study: Production Scheduling},label={fig:data-case-study},caption={Selected task fields and one complete case input from a production scheduling episode. The hidden rule summary and reference output are evaluation-side annotations. Field descriptions and ancillary rule parameters are omitted where indicated. The same input appears as case\_059 in Figure~\ref{fig:error-case-study}.}]
[Task]
At each short-horizon scheduling checkpoint, a production planning
supervisor decides which sequencing intervention to apply to a released
work order over the next production window.

[Actions]
action_0: Keep current sequence
action_1: Resequence within assigned line
action_2: Authorize overtime capacity
action_3: Reroute to alternate capacity

[Field catalog, excerpt]
"remaining_labor_hours": {
  "description": "The standard qualified labor-hours required by the unfinished routing operations of this work order.",
  "type": "decimal", "unit": "labor-hours", "minimum": 0, "maximum": 240
}
"line_throughput_trend": {
  "description": "The slope of hourly good-unit throughput rates on the assigned line across the trailing 8 operating-hour observation window.",
  "type": "decimal", "unit": "units per hour squared", "minimum": -16, "maximum": 16
}
...

[Case input from batch_08]
{"context_id": "case_059", "raw_values": {
  "line_throughput_trend": -10.63,
  "remaining_labor_hours": 67.17,
  "expected_machine_recovery_hours": 2.32,
  "forecast_machine_unavailability_hours": 1.52,
  "required_changeover_hours": 0.16,
  "changeovers_ahead_count": 0,
  "historical_line_downtime_rate": 0.12,
  "downstream_blocked_order_count": 1
}}

[Hidden rules, excerpt; + / - denote factor directions]
Family: additive_aggregation. Change occurs after batch_07.
Before: remaining_labor_hours (+), line_throughput_trend (-),
        expected_machine_recovery_hours (+)
After:  remaining_labor_hours (+), line_throughput_trend (+),
        forecast_machine_unavailability_hours (+)
...

[Reference output, hidden from the agent]
{"optimal_action_id": "action_1", "utilities": {
  "action_0": 0.14, "action_1": 0.5,
  "action_2": -0.31, "action_3": -0.33
}}
\end{template*}


\begin{template*}[title={Error Case Study: A Missed Direction Reversal},label={fig:error-case-study},caption={Trajectory excerpts for GPT-6 Sol under Default on the episode in Figure~\ref{fig:data-case-study}. The paired probes differ only in throughput trend. The agent chooses action\_3 for case\_059 and later reports the correct replacement factor but the old throughput direction. Other probes, cases, feedback fields, and intervening interactions are omitted with ellipses.}]
[Agent factor assessment before the change]
{"relevant_fields": ["line_throughput_trend", "remaining_labor_hours",
                     "expected_machine_recovery_hours"],
 "field_directions": [
  {"field_id": "line_throughput_trend", "direction": "negative"},
  {"field_id": "remaining_labor_hours", "direction": "positive"},
  {"field_id": "expected_machine_recovery_hours", "direction": "positive"}
 ]}
...
[Retained base probe]
probe_059: remaining_labor_hours = 8; all other input fields = 0.
[Agent -> Probe Batch after batch_07, excerpt]
{"tool": "probe_batch", "arguments": {"probes": [
  {"base_context_id": "probe_059", "changes": {
    "line_throughput_trend": -1, "remaining_labor_hours": 40,
    "expected_machine_recovery_hours": 2,
    "forecast_machine_unavailability_hours": 2}},
  {"base_context_id": "probe_059", "changes": {
    "line_throughput_trend": 1, "remaining_labor_hours": 40,
    "expected_machine_recovery_hours": 2,
    "forecast_machine_unavailability_hours": 2}},
  ...
]}}
[Environment -> Agent; case_059 input is shown in the data case]
{"results": [
  {"context_id": "probe_103", "utilities": {
    "action_0": -0.26, "action_1": 0.67,
    "action_2": -0.08, "action_3": -0.33}, ...},
  {"context_id": "probe_104", "utilities": {
    "action_0": -0.3, "action_1": 0.51,
    "action_2": 0.12, "action_3": -0.33}, ...},
  ...
], "pending_action_batch": {"batch_id": "batch_08", "cases": [
  ...,
  {"context_id": "case_059", "raw_values": {...}},
  ...
]}, ...}
[Agent -> Submit Action Batch, excerpt]
{"tool": "submit_action_batch", "arguments": {
  "batch_id": "batch_08", "actions": {..., "case_059": "action_3", ...}
}}
[Environment -> Agent, excerpt]
{"batch_id": "batch_08", "results": [
  ..., {"context_id": "case_059", "action_id": "action_3", "utility": -0.33}, ...
], ...}
...
[Agent factor assessment at episode completion, after batch_12]
{"relevant_fields": ["line_throughput_trend", "remaining_labor_hours",
                     "forecast_machine_unavailability_hours"],
 "field_directions": [
  {"field_id": "line_throughput_trend", "direction": "negative"},
  {"field_id": "remaining_labor_hours", "direction": "positive"},
  {"field_id": "forecast_machine_unavailability_hours", "direction": "positive"}
 ]}
\end{template*}


\begin{template*}[title={Field Discovery Prompt},label={fig:prompt-field-discovery},caption={System prompt for discovering decision-relevant observable fields and their intervention directions.},listing options={basicstyle=\ttfamily\footnotesize,breaklines=true,columns=fullflexible,keepspaces=true,literate={—}{\textemdash}1}]
You build the complete practical inventory of numeric measurements that support decision making in the supplied operational scenario. Return every realistic measurement that can realistically inform intervention tendency. Explore current state, exposure, workload, time, quality, exceptions, risk, recovery, forecasts, trends, completed historical evidence, and decision-relevant subgroups when they answer materially different operational questions; Give a diverse set while preventing same or synonymous fields.

Include an action-neutral numeric measurement when it describes or predicts the governed condition's expected loss, deterioration, urgency, or autonomous recovery within the decision horizon. It may describe the governed item or a causally relevant route, cohort, queue, environment, or system. Exclude pure regime selectors, action authorization or execution capability, consequences of choosing a stronger action, action-conditional forecasts, and the unresolved governed item's post-decision outcome. 

For every field, output only a stable field_id, concise display_name, integer or decimal type, exact unit, one concise description, and pressure_direction. The description must identify exactly one decision-evidence quantity and explain what the reported numeric value represents. To avoid ambiguity, you should make necessary clarifications as follows: When needed to distinguish the quantity, name the operands and their order for a difference; the numerator, denominator, and time basis for a fraction or rate; the target event and horizon for a probability or forecast; and the governed population, observation window, and aggregation for a cohort statistic. For field whose value may be not existing under certain circumstances, state it as undefined or null, but do not assign a value that may cause ambiguity. For instance, for field "preparation_time", when there is no preparation at all, the value does not exist, state it as null/undefined instead of assigning 0, else there could be ambiguity for preparation_time=0 (do not know whether there is preparation but finished immediately or there is no preparation).

For any one fixed surrounding decision condition, every retained field must have a well-defined monotonic direction across its full defined legal numeric domain (do not need to consider situations where the field is undefined or unavailable). The direction must remain stable while that condition is fixed (but it may be different under another fixed condition). Across the field's entire legal domain, increasing the value must consistently support either the same or stronger intervention, or the same or weaker intervention; the direction must never switch within that domain. Omit a field whose middle values are preferable while both lower and higher extremes require stronger intervention, or whose direction otherwise changes merely because its own value crosses a threshold.

While each field should have a clear intervention tendency within one certain condition, the direction are allowed to flip across different regime. During discovery, actively and broadly look for fields whose entire monotonic direction could genuinely reverse after a concrete, scenario-plausible event or condition change. For instance, "machine uptime minutes" normally means more availability and less intervention pressure, but during a mandated preventive-shutdown window the same high uptime means the required outage was skipped, which requests for more intensive intervention. So do not limit your discover only to those fields who obviously point towards an adverse or intervention needs, deliberately brainstorm for independently useful, action-neutral direct measurements, such as direct operational levels, signed flows or rates, progress or completion, throughput, uptime or availability, elapsed or remaining time, neutral forecasts or trends, whose pressure direction is clear under the ordinary condition but may reverse after a separate external event or condition change.

Assign pressure_direction according to the field's ordinary interpretation in the recurring decision, if you consider a field to be reversible, then it may have different direction under different circumstances, assign the direction for the most common situation you think: use 1 when increasing values monotonically support stronger intervention and -1 when increasing values monotonically support weaker intervention. Notice that we require the pressure_direction to be applied to the whole domain under one specific circumstances. A plausible conditional reversal does not make the base pressure_direction ambiguous.

Also consider distinct measurements that could both serve the same decision role when a concrete workflow event changes which measurement is valid evidence, where they could potentially replace each other to form the decision rule.

Avoid obvious semantic aliases, exact complements, and deterministic transforms. That is to say, if 2 or more fields refer to exactly the same fact, or they are completly dependent on each other, for instance, mass and weight, speed and velocity, then you should reserve only one of them, eliminate the redundant fields. Each field in the final output set should be distinguishable from each other.

Self check before you finally output the field inventory: 
- You should have brainstormed as diverse as possible to give every realistic measurement that could inform intervention tendency under the scenerio;
- You should have assigned a stable field_id, concise display_name, integer or decimal type, exact unit, one complete description, and pressure_direction to each field;
- No field should be an alias or fully dependent on others, redundant or non-distinguishable ones have been eliminated.
\end{template*}

\begin{template*}[title={Field Semantics Prompt},label={fig:prompt-field-semantics},caption={System prompt for assigning field significance, eligible rule families, and direction-reversal semantics.},listing options={basicstyle=\ttfamily\footnotesize,breaklines=true,columns=fullflexible,keepspaces=true,literate={—}{\textemdash}1}]
Your job is to analyze the input fields according to their description and other given information, then assign effect_strength, eligible_families and reversal semantics to each of them. You should consider the complete scenario, action ladder, field inventory, and fixed rule-family meaning.

Output exactly as required, where field_id is used to refer to the given field, so do not edit or change it, otherwise you would be assigning values/texts to a wrong field.

Effect_strength is for the field's quantified decision relevance relative to the complete catalog, which reflects how significant the field is in affecting the final decision making compared to other fields. Assign a value in [0.1, 1.0] for each field. Effect_strength regards the relative significance, that is to say, fields with effect_strength 0.1 only means that it is less severe than other fields, it does not mean the field is not important, so use the full range meaningfully to make each field more distinguishable.

Eligible_families regards whether the measurement genuinely supports: graded contribution for additive aggregation, a meaningful joint gate for conjunctive gating, or an independently decisive severe value for disjunctive sufficiency. Select the supported roles for each field after fully analyzing the field's role and the three supplied family meanings.

Is_reversal examines whether a field's impact on the intervention tendency could reverse across different conditions. For instance, "machine uptime minutes" normally means more availability and less intervention pressure, but during a mandated preventive-shutdown window the same high uptime means the required outage was skipped, which requests for more intensive intervention — the operating rule changed, the measurement did not. Mark is_reversal=true for such kind of fields, when there exists reasonable event that could genuinely reverses what a higher value of this same unchanged measurement implies for intervention on the same action ladder. Notice that the event may be external changes; for example, in cybersecurity operations, higher outbound traffic may normally increase exfiltration concern, while during an approved backup window the same measurement can instead indicate expected backup progress and reduce intervention pressure. Also, the magnitudes of the two tendencies do not need to be similar or symmetric; any genuine nonzero opposite monotonic tendency qualifies as a reversal. You must fully brainstorm the existence of such kind of reversal event before you finally assign a "false".

However, be aware that reverse means a flip in direction on intervention tendency; a change in action cost, ease, authority, or execution capability is not a reversal. A field that changing into meaningless, undefined or losing its impact after the trigger being activated should also not be considered as reversal. For instance, a field A may have pressure direction 1 (positive impact on intervention tendency), but after the event happens, field A may no longer influence the urgency of taking immediate intervention, this kind of changes are not considered as reverse. 

For each field that you assign "is_reversal=true", you should further consider the rationale and trigger, fill all four event-fields with one concrete but consise sentence:
- trigger: the concrete scenario-plausible operational event or conditional change that flips the tendency; Notice that trigger should be independent to the action decision and the field's actual value, it must not pre-decide the action or field value to make it affect the pressure direction.
- semantics_before: how a higher value of this unchanged measurement maps to intervention tendency in the original condition, be careful to confirm that this should agree to the given pressure direction value;
- semantics_after: how the same higher observation maps to the opposite intervention tendency after the triggerred event;
- reversal_reason: why the regime genuinely reverses the causal meaning of the same measurement.
For is_reversal=false, set these four fields to empty strings and use reversal_rationale to explain what conditions and event were considered and tested and why the tendency should remains stable across different conditions. You should write details of what conditions and events have been tested, do not casaully or briefly skip the brainstorm just because the reverse is not obvious. For true, set reversal_rationale empty.

Before returning, check the following: all fields' effect_strength, eligible_family, reversal semantics has been fully assigned values; effect strengths must be globally comparable; family roles must follow the fixed mathematical meanings; and every reversal account has been fully analyzed.
\end{template*}

\begin{template*}[title={Field Parameter Calibration Prompt},label={fig:prompt-field-parameters},caption={System prompt for specifying realistic field ranges and Beta population parameters.},listing options={basicstyle=\ttfamily\footnotesize,breaklines=true,columns=fullflexible,keepspaces=true,literate={—}{\textemdash}1}]
You calibrate realistic synthetic numeric priors for frozen operational measurements whose decision semantics are already accepted. Jointly author legal range, typical range, and low/mixed/high Beta populations for every assigned field. Do not change field identity, measurement semantics, relative strength, family eligibility, or reversal properties.

Author numerical calibration for every assigned field exactly once. Use the complete catalog to keep all batches grounded in the same ordinary, mixed, and stressed scenario populations.

For each assigned field:
1. Set a credible legal minimum and maximum in its frozen unit, with a narrower typical range covering ordinary operation. Legal and typical ranges are different concepts. Choose the legal range around values realistically observable across the field's common and stressed scenario populations, considering the intended profile means; do not stretch it to a merely theoretical extreme that makes common values occupy a tiny fraction of the range. Explicitly verify `minimum <= typical_range[0] <= typical_range[1] <= maximum`.
2. Define low, mixed, and high intervention-tendency populations over normalized x=(v-minimum)/(maximum-minimum) using Beta(alpha,beta). Each alpha and beta must be strictly greater than 0, and each concentration alpha+beta must be from 4 through 12 inclusive. The normalized mean is m=alpha/(alpha+beta), and the mapped raw mean is minimum+m*(maximum-minimum).
3. For pressure_direction=1 require m_low <= m_mixed <= m_high; for pressure_direction=-1 require m_low >= m_mixed >= m_high. The mapped raw means for both low and mixed must fall inside the authored typical range. The high center may leave the typical range when it remains a plausible stressed population, but should approach a legal extreme only when the scenario supports such a population.
4. Choose concentrations alpha+beta that express plausible within-profile variation. Shared shapes are valid only when field semantics justify them; do not copy one universal Beta template.
5. In parameter_basis, summarize the legal/typical range basis and the population centers and variance. Treat emitted ranges and Beta pairs as numerical authority; do not restate calculated means as prose numbers.

You should calculate normalized and mapped raw means for all three profiles of each field. Verify their order against the frozen pressure_direction. Write down the required inequality internally before choosing shapes: direction 1 means low <= mixed <= high; direction -1 means low >= mixed >= high. Verify both low and mixed mapped means lie inside the authored typical range, not merely inside the legal range. Then verify that parameter_basis describes those exact populations without reversing their meanings. Do not casually assign a value without verification.
\end{template*}

\begin{template*}[title={Lexicographic Order Prompt},label={fig:prompt-lexicographic-orders},caption={System prompt for deriving admissible three-field priority orders for lexicographic decision rules.},listing options={basicstyle=\ttfamily\footnotesize,breaklines=true,columns=fullflexible,keepspaces=true,literate={—}{\textemdash}1}]
Your job is to discover as many usable three-field lexicographic groups and to author the defensible first-to-last priority rules for those groups. Treat every supplied field as a candidate, but do not force a field into an invalid group merely to increase coverage or count.

A lexicographic group is a set of three distinct fields that can serve as successive decision cues for the same scenario, decision objective, and action ladder. Each field must have a realistic, nonempty region in which its value can independently settle an action or intervention level without being numerically combined with the other two; Each field must also have a meaningful indifference or inconclusive region in which a later cue can still provide residual decision evidence. In this way, the three fields [A,B,C] could form a priority rule: If A satisfy a condition, then it directly decides the final decision, no matter what values B and C take; Else if B satisfy a condition, then it decides the final decision regardless of C; Else we make the final decision based on the value of C. The three fields should not be aliases, deterministic restatements, fields whose decision meaning depends on jointly calculating another member, mutually exclusive measurements with no realistic common decision condition, and sets for which no member can leave a meaningful unresolved case to the next cue. A valid group needs at least one coherent ordering, but its priority need not be inherent, universal, or the only defensible policy choice.

For example, [current_collateral_shortfall_usd, settlement_failure_probability, minutes_to_settlement_cutoff] can form a valid lexicographic group: a material collateral shortfall may independently settle the intervention level; when the shortfall is within its indifference region, settlement-failure probability provides a distinct risk cue; and when both are inconclusive, time to cutoff supplies the final urgency cue. Likewise, [vehicle_range_remaining_km, driver_duty_time_remaining_minutes, shift_time_remaining_minutes] can form a valid group: critically insufficient vehicle range may require recovery on physical-feasibility grounds; when range is not decisive, insufficient legal duty time may independently require reassignment; and when neither constraint settles the case, remaining shift time provides the final operational-horizon cue. These examples illustrate how distinct fields can successively and independently determine an action without being jointly calculated. They are not fixed patterns: consider other decision roles and retain only substantively justified orderings rather than automatically emitting every permutation.

Your task is to find as many distinct valid three-field groups as the catalog meaningfully supports, then determine which exact orderings of each group have a scenario-grounded rationale. In an ordered rule [first, second, third], the first field settles the decision whenever it is materially decisive and otherwise passes the case to the second; the second does the same for the cases left unresolved by the first; the third supplies the final cue when both earlier fields are inconclusive. Notice that we do not require the ordering to be concrete or very obvious, because the order is a rule to be established, we only require the group to be reasonable and not counterintuitive. Fields may participate in many overlapping groups, and the same unordered group may support more than one ordering. Do not partition the catalog into disjoint triples, stop after a convenient number, or emit all six permutations by default without fully analyzing. Do not set up groups purely based on the fields' name, you should think based on their description and actual role.

You can discover the lexicographic groups with the following thinking strategy:
1. Read every field's description and identify the decision role it can independently play, such as current state or severity, consequence magnitude, likelihood, urgency or time horizon, trend or persistence, recoverability, evidence reliability, or execution feasibility. Identify both the readings that can settle an action and the readings that would leave the decision unresolved.
2. Build overlapping candidate neighborhoods around shared decision concerns and complementary decision roles. A field may belong to several neighborhoods. Do not require all three fields to measure the same concept, and do not treat disagreement between their indications as incompatibility.
3. Within each neighborhood, identify strong anchor pairs: two nonredundant fields that can each independently settle some cases and can sensibly be consulted in succession without requiring a joint calculation.
4. Expand each anchor pair with every substantively distinct third field that can resolve cases left inconclusive by the pair. Revisit a strong pair with alternative third cues instead of accepting only the first plausible completion.
5. Revisit fields that are absent or rarely used. Search for additional neighborhoods, anchor pairs, and third-cue roles that connect them to the rest of the catalog, but do not force every field to belong to a lexicographic group
6. Deduplicate the resulting unordered groups. For each retained group, compare the plausible first, second, and third assignments and keep only the orderings whose three positions have distinct, concrete policy roles. Similar wording with the positions permuted is not a sufficient rationale.
7. Make a final coverage sweep across decision roles, semantic neighborhoods, reused anchor pairs, and underused fields so that obvious valid groups are not omitted.

For every retained ordered rule, output exactly three distinct supplied field IDs in first-to-last order, a stable and unique order_id that distinguishes alternative orders, and one concise rationale tied to that exact sequence. Return the supported ordered triples in the required flat output format; do not output intermediate neighborhoods, anchor pairs, unordered-group records, quotas, or unsupported permutations.
\end{template*}

\begin{template*}[title={Cross-Field Constraint Prompt},label={fig:prompt-field-constraints},caption={System prompt for constructing executable cross-field feasibility constraints.},listing options={basicstyle=\ttfamily\footnotesize,breaklines=true,columns=fullflexible,keepspaces=true,literate={—}{\textemdash}1}]
Your job is to compare against different fields and author executable cross-field feasibility constraints according to their definition.

First identify fields connected by a strong definitional, accounting, population-containment, or physical-capacity relationship. Emit a constraint only when their jointly observed numeric values must satisfy that relation, the constraint should be some hard restriction to avoid value combinations that is obviously impossible in real world. For instance, the combination "total_number_of_objects = 30" and "number_of_type_A_object = 40" may be impossible if type A object belongs to total objects, a union cannot be less than/smaller than its part. Ordinary correlation, causal expectation, common co-movement, and typical behavior are not invariants; an empty list is valid.

Evaluate a constraint only over records where every field it references has a numeric value. Only consider cross field relations and constraints, do not encode availability, nullness, missingness, benign non-applicability, or single-field legal ranges in the expression, and never use a field's ordinary range bound as a guard for a cross-field relation. Prefer one direct comparison whenever it states the invariant. When a genuine conditional invariant is needed, use `implies` explicitly instead of using an equivalent expression of `or`.

Express each invariant with the finite structured expression grammar. A numeric atom is {"field": FIELD_ID}. A numeric term is an atom, {"const": NUMBER}, or one arithmetic node over atoms; `add` and `mul` take two to eight atoms, while `sub` and `div` take exactly two. A numeric expression is a term or one arithmetic node over terms with the same arities. A comparison uses `le|lt|ge|gt|eq` with numeric-expression `left` and `right`. A Boolean clause is a comparison, `not` of one comparison, or `and|or` over two to eight comparisons. A complete Boolean expression is a clause, `not` of one clause, `and|or` over two to eight clauses, or `implies` whose `if` and `then` are clauses. Do not nest beyond these layers. Split independent conjuncts into separate constraints instead of nesting repeated `and` nodes. All arithmetic and comparisons must be dimensionally valid. Operands of `add` and `sub` must have compatible dimensions, and the two sides of every comparison must have compatible dimensions after any explicit exact unit conversion. `mul` and `div` may be used only when their resulting dimensions make the comparison valid. Never directly compare a count with a rate or duration.

A constraint can be set among more than 2 fields, for each candidate, verify only the actual definitions to establish the constraint, do not judge merely from the field name. If any valid joint numeric assignment can violate the proposed relation, omit it. Notice that independent constraints should be established separately, thus long constraints are used only when necessary, do not combine several short constraints into a whole long expression.

Before returning, compare all retained constraints. If a constraint contains could be devide into two or more independent constraints, then split them. If two constraints impose the same restriction after commutative reordering or algebraic simplification, emit only the clearer, shallower one. Do not treat partial overlap or one-way implication as duplication.

Give each retained constraint a stable id and one concise description stating why the numeric restriction is necessary. Every constraint must have a non-empty ID that is unique within feasibility_constraints. Never reuse an ID for two constraints, even when their descriptions or expressions differ. Before returning, confirm that every field reference is one supplied ID and evaluate the exact AST on its boundary cases to ensure it states the same restriction as the description. Similar names alone do not establish equality, complementarity, containment, or interchangeability.
\end{template*}

\begin{template*}[title={Field Replacement Prompt},label={fig:prompt-field-replacements},caption={System prompt for specifying event-grounded field replacements and their directional transitions.},listing options={basicstyle=\ttfamily\footnotesize,breaklines=true,columns=fullflexible,keepspaces=true,literate={—}{\textemdash}1}]
Your job is to inspect the complete frozen field inventory and find every two-field replacement supported by a coherent event. Do not preselect by similar names, identical units, or an assumption that replacements must naturally occupy exactly the same status.

A replacement means that a concrete scenario-plausible event or condition change can make one field reasonably take the other field's place in the same rule slot. The two measurements do not need to be identical, interchangeable in every situation, or equally authoritative before and after the event. They need a defensible reason why the incoming field becomes the more appropriate cue when the outgoing field stops occupying that slot. Retain a pair only when both fields have the same frozen `pressure_direction`, so the handoff preserves the slot's intervention meaning. Different types, units, and raw constructs are allowed and are not rejection criteria.

Consider every field pair in both directions. Keep a pair whenever you can explain a concrete handoff event that:
- makes the outgoing field unavailable, stale, out of scope, less representative, or no longer the preferred evidence for that rule position; and
- makes the incoming field available and relevant enough to take that position without contradicting the intervention meaning of the outgoing field.

Do not require the incoming field to reproduce exactly the same quantity, population, horizon, or evidentiary status. Those differences may be the reason replacement becomes useful. Exclude a pair only when the frozen meanings have incompatible intervention implications in the claimed circumstances, the two fields must necessarily be used together rather than one taking the other's slot, or no coherent event can justify the handoff.

Classify the relationship according to the actual rationale:
- same_metric: alternate sources report the same construct with the same frozen pressure direction;
- different_horizon: compatible measurements describe the same construct at different horizons or observation cutoffs with the same frozen pressure direction;
- same_source: distinct views from one governing system can replace one another after a workflow or authority change;
- proxy_substitution: different constructs can occupy the same rule position under different conditions because each is a reasonable proxy for the decision concern then relevant. Raw types and units may differ, but the frozen pressure direction must remain the same.

For each retained group, output a unique `group_id`; the two `members`; the honest `relation_type`; a concise `decision_role` naming their shared rule function; a pair-specific `rationale` explaining the shared decision concern, the meaningful difference between the measurements, and why that difference still permits the event-specific handoff; a concise `source_contrast`; exactly one `authority_by_field` entry for each member; a `coupling_strength` from 0 through 0.6 representing expected normalized dependence; the `effective_scope`; and the two directed `transitions` required below. `effective_scope` must describe the governed business objects or population, workflow or decision boundary, and relevant time or authority boundary within which the handoff applies. It is not a list of rule families or field eligibility labels. Do not let `rationale` merely restate that one field becomes unavailable and the other becomes available.

Author exactly two directed transitions, one in each direction. Each direction needs its own plausible trigger, a reason the outgoing field relinquishes the slot, a reason the incoming field can take it, an expected duration, and an ending condition. The two directions may rely on different events and need not be symmetric or equally common. A transition is invalid only when it removes the outgoing field but provides no positive reason for using the incoming field.

Return all supported groups found by the complete pairwise sweep. Fields may participate in multiple groups. `capacity_findings` should briefly record the search and concrete near-candidates that truly lacked any coherent handoff; an empty replacement list requires specific evidence that the strongest candidate pairs were considered.
\end{template*}

\begin{template*}[title={Candidate Semantic Review Prompt},label={fig:prompt-candidate-semantic-review},caption={System prompt for reviewing candidate episode field configurations, rule changes, and distractors.},listing options={basicstyle=\ttfamily\footnotesize,breaklines=true,columns=fullflexible,keepspaces=true,literate={—}{\textemdash}1}]
You audit candidate decision episodes for real-world coherence. Treat every
candidate independently. Judge whether the declared rule roles and all eight
visible measurements can coexist for the same governed entity, decision time,
and horizon. For replacement and reversal candidates, assess only the declared
operation; do not invent an event, rationale, or additional rule change.

Use reject_core when the relevant field roles or declared operation are
incoherent. Use reject_distractors only when the core is coherent but the
distractor set cannot coexist with it or leaks a post-decision outcome. Use
accept when the core and complete visible field set are coherent. Do not propose
replacement fields. A candidate with prior_core_review has already passed its
core review: judge only its newly drawn distractors and return only
reject_distractors or accept. Return only the strict JSON object requested by
the response schema.
\end{template*}

\begin{template*}[title={Dataset-Level Quality Review Prompt},label={fig:prompt-final-dataset-review},caption={System prompt for scoring field realism, rule-change realism, and distractor quality in dynamic episodes.},listing options={basicstyle=\ttfamily\footnotesize,breaklines=true,columns=fullflexible,keepspaces=true,literate={—}{\textemdash}1}]
You are the final semantic quality reviewer for a dynamic decision-rule dataset.
Score every supplied episode independently on all three dimensions from 0.0 to
5.0 in increments of 0.1. Use the full scale: 5 is exemplary, 4 is strong, 3 is
acceptable with visible weaknesses, 2 is substantially weak, 1 has a severe
problem, and 0 is incoherent or contradictory.

field_realism measures whether the eight fields can naturally be available for
the stated decision, governed entity, time, and horizon, and whether the field
combination fits the scenario.

rule_change_realism measures whether the stated trigger_event and change_reason
plausibly produce the exact replacement and direction reversal shown by the
pre/post core fields and directions. A compound change must follow from one
coherent observable operational event rather than two unrelated events.

distractor_quality measures whether distractors are plausible, superficially
relevant context without directly encoding the action, leaking the outcome, or
semantically subsuming the rule core. Do not penalize a distractor merely for
being correlated with the decision or occasionally seeming important in a
randomly sampled case; penalize it only when its definition clearly dominates,
duplicates, or bypasses the intended core mechanism.

Give exactly one concise sentence (at most 240 characters) explaining each
dimension score. Return every episode_id exactly once using only the requested
JSON schema.
\end{template*}

\begin{template*}[title={Initial Prompt},label={fig:inference-prompt},caption={The initial prompt under Default. The system message is identical across episodes, and angle brackets mark the episode-specific role, action, and field descriptions. The first user message closes with the initial Current state card.},listing options={basicstyle=\ttfamily\footnotesize,breaklines=true,columns=fullflexible,keepspaces=true,literate={—}{\textemdash}1}]
===== System message =====

# Role

<role: one sentence assigning the domain decision task>

Choose one of the four actions for every case. Your objective is to maximize
Cumulative Utility — the total utility earned by submitted actions.

Probes reveal information about the hidden rule. When choosing probe cases,
avoid redundant probes, such as repeating the same case or using cases whose
field values are nearly identical. The decision rule may or may not change
during the episode.
You are not told if or when it changes; a change can only happen right after
you complete an action batch.

# Actions

- **<action display name>** (`action_0`): <what this action does in the domain>
- **<action display name>** (`action_1`): <what this action does in the domain>
- **<action display name>** (`action_2`): <what this action does in the domain>
- **<action display name>** (`action_3`): <what this action does in the domain>

# Field relevance

Some listed fields are distractors. Only a subset affects the hidden rule and
utility; changing the other fields cannot change utility. The relevant subset
is hidden, so infer it from probe results and action feedback. It may change if
the decision rule changes during the episode.

# Rule family

The hidden rule uses exactly one of these four families:

- `additive_aggregation`: combine weighted contributions from relevant fields.
- `conjunctive_gating`: all conditions matter; the weakest condition controls.
- `disjunctive_sufficiency`: any sufficient condition can control; the strongest
  condition controls.
- `lexicographic_priority`: inspect fields in priority order; the first
  decisive field controls, otherwise fall back to the last field.

Infer the family from probe results and action feedback. The family name and
field roles are not given directly.

# Utility scale

Each submitted action receives a per-case utility from a fixed four-action
mapping. Utilities are not normalized to `[-1, 1]`: across the full score
range, an individual utility is approximately between `-0.33` and `0.98`.
The theoretical lower and upper limits are about `-0.333` and `0.976`.
Positive utility is better, negative utility is worse, and `0` is neutral;
the four utilities for one case sum to approximately `0`. Abstaining also
receives utility `0`, but is counted as incorrect. Maximize the sum of
utilities across submitted actions.

# Reply format

Every reply is exactly one JSON object with two keys, `tool` and `arguments`,
and nothing else — no prose, Markdown, rationale, or extra keys. Example:

```json
{"tool":"request_next_batch","arguments":{}}
```

Use only a tool marked available in the latest `tool_usability`; an accepted
call is final.

# Tools

## probe_batch — look at example cases before deciding

Each probe shows one case and the utility of all four actions on it. Probe
results are information only and earn no utility; action feedback exposes only
the per-case utility of the actions you submitted.

Two probe forms. A fresh case — give a value for every field in
`raw_field_schema`:

```json
{"tool":"probe_batch","arguments":{"probes":[{"values":{"<field_id>":1}}]}}
```

A variation of a known case — start from a context ID returned by probe
feedback and edit some of its fields:

```json
{"tool":"probe_batch","arguments":{"probes":[{"base_context_id":"<context_id>","changes":{}}]}}
```

Limits: one `probe_batch` call per batch window, at most
16 probes per call. Respect field types, feasibility
constraints, and two-decimal precision for decimals.
Probe policy: submit at least one valid probe when you choose `probe_batch`.

## request_next_batch — reveal the next batch without probing

```json
{"tool":"request_next_batch","arguments":{}}
```

## submit_action_batch — answer the pending cases

```json
{"tool":"submit_action_batch","arguments":{"batch_id":"batch_01","actions":{"<context_id>":"action_1"}}}
```

`pending_action_batch` lists 8 context IDs; give at most one
action for each. You may omit IDs to abstain — abstained cases receive
utility 0 and count as incorrect — and `actions` may even be empty.

## calculator — optional arithmetic helper

```json
{"tool":"calculator","arguments":{"expressions":["total = (12.5+3)*4","sqrt(total)"]}}
```

Pure arithmetic: numbers, operators, parentheses, and the functions `abs`,
`sqrt`, `min`, `max`, `sum`, `mean`, `round`, `floor`, `ceil`, `log`, `log10`,
`exp`, `pow`. `name = expression` assigns a variable and returns its value,
and within one call you may reference a variable before its assignment
appears. Variables are call-local: to reuse a value in a later call, copy the
number into the new expression. After a calculator call, use a different tool
before calculating again — two calculator calls in a row are not allowed.

# Batch flow

1. Batch 1 is hidden. Call `probe_batch` or `request_next_batch` to reveal it.
2. When `pending_action_batch` exists, call `submit_action_batch`; otherwise
   choose `probe_batch` or `request_next_batch`.
3. After each submitted batch except the last, again choose `probe_batch` or
   `request_next_batch` before the next batch.
4. A rejected call gets one repair opportunity. If the second call is invalid,
   a hidden batch is revealed by skipping probe; a pending action batch fails
   with every case unanswered at utility 0, and the environment advances to
   the next window.
5. Use the latest state card as authoritative. Do not infer hidden rules or
   metadata from identifiers.

# State and history

After you complete an action batch, the environment rewrites retained history
before the next request: the revealed cases, your `submit_action_batch` call,
and your action-solving messages are removed, and an environment-authored
outcome replaces them. The next message opens like this:

```
Environment result:
{
  "message": "The environment accepted action batch 3. Your selected actions and action-solving messages are hidden from later agent context.",
  "batch_number": 3,
  "results": [{"context_id": "case_017", "utility": 0.59}, {"context_id": "case_018", "utility": -0.31}],
  "batch_utility": 1.33,
  "cumulative_utility": 5.62,
  "cumulative_probes": 8
}
```

This outcome means the named batch was just hidden, and it is your only record
of that round. Treat it as a scoreboard for your induced rule: each utility is
the payoff of the action you chose, so high utilities confirm the rule and low
or negative ones flag mismatches. If a batch scores markedly worse than
earlier batches under the same hypothesis, your rule is wrong or the rule
changed after the previous batch — consider whether additional probing would
help.

Only context IDs returned by probe feedback may be used as bases for
incremental probes; use complete values for a new probe.

The initial user message contains field descriptions and a `Current state` card.
Later messages contain an `Environment result` followed by the latest card.
Use the latest card for tool usability, `cumulative_utility`,
`cumulative_probes`, pending cases, and remaining action batches; use
conversation history for earlier probe/action evidence.

===== First user message =====

Input fields:
Every field below is a numeric probe input and is required in a full `values` probe. Use JSON numbers only: `integer` requires a whole number and `decimal` accepts an integer or decimal number. `null`, strings, and booleans are invalid, even when a field description discusses an unavailable real-world observation.
- **<field display name>** (`<field_id>`, type: <integer or decimal>, unit: <unit>): <field description>
... one bullet per field ...

Current state:
{
  "cumulative_utility": 0,
  "cumulative_probes": 0,
  "remaining_action_batches": 12,
  "pending_action_batch": null,
  "tool_usability": [
    {
      "tool": "probe_batch",
      "available": true,
      "reason": null
    },
    {
      "tool": "request_next_batch",
      "available": true,
      "reason": null
    },
    {
      "tool": "submit_action_batch",
      "available": false,
      "reason": "no action batch is pending; reveal one with probe_batch or request_next_batch first"
    },
    {
      "tool": "calculator",
      "available": true,
      "reason": null
    }
  ]
}
\end{template*}

\begin{template*}[title={Probe Batch Feedback},label={fig:prompt-probe-batch},caption={Environment feedback to a Probe Batch call under Default. Each probe entry reports the utility of all four actions on the probed case, and the updated state card reveals the pending action batch. Entries that repeat a displayed structure are elided.}]
Environment result:
{
  "results": [
    {
      "context_id": "probe_001",
      "probe_mode": "new_context",
      "utilities": {
        "action_0": 0.94,
        "action_1": -0.27,
        "action_2": -0.33,
        "action_3": -0.33
      }
    },
    <one entry per probe case; an existing-context probe also carries its "base_context_id">
  ]
}

Current state:
{
  "cumulative_utility": 0,
  "cumulative_probes": 16,
  "remaining_action_batches": 12,
  "pending_action_batch": {
    "batch_id": "batch_01",
    "batch_number": 1,
    "cases": [
      {
        "context_id": "case_001",
        "raw_values": {
          "<field_id>": <value>,
          ...
        }
      },
      <one entry per case in the batch>
    ]
  },
  "tool_usability": [
    <one availability entry per tool, as in the initial state card>
  ]
}
\end{template*}

\begin{template*}[title={Request Next Batch Feedback},label={fig:prompt-request-next-batch},caption={Environment feedback to a Request Next Batch call under Default. Revealing a batch without probing produces no separate result payload, so the reply consists of the updated state card alone.}]
Current state:
{
  "cumulative_utility": 1.92,
  "cumulative_probes": 16,
  "remaining_action_batches": 11,
  "pending_action_batch": {
    "batch_id": "batch_02",
    "batch_number": 2,
    "cases": [
      {
        "context_id": "case_009",
        "raw_values": {
          "<field_id>": <value>,
          ...
        }
      },
      <one entry per case in the batch>
    ]
  },
  "tool_usability": [
    <one availability entry per tool, as in the initial state card>
  ]
}
\end{template*}

\begin{template*}[title={Submit Action Batch Feedback},label={fig:prompt-submit-action-batch},caption={Environment feedback to a Submit Action Batch call under Default. The result reports the per-case utility of the submitted actions, and the state card advances to the next batch window.}]
Environment result:
{
  "batch_id": "batch_01",
  "results": [
    {
      "context_id": "case_001",
      "action_id": "action_2",
      "utility": 0.72
    },
    <one entry per case in the batch>
  ],
  "batch_utility": 1.92,
  "cumulative_utility": 1.92,
  "cumulative_probes": 16,
  "tool_usability": [
    <one availability entry per tool>
  ]
}

Current state:
{
  "cumulative_utility": 1.92,
  "cumulative_probes": 16,
  "remaining_action_batches": 11,
  "pending_action_batch": null,
  "tool_usability": [
    <one availability entry per tool, as in the initial state card>
  ]
}
\end{template*}

\begin{template*}[title={Calculator Feedback},label={fig:prompt-calculator},caption={Environment feedback to a Calculator call under Default. Results follow the submission order of the expressions, and assignments additionally report the bound variable.}]
Environment result:
{
  "results": [
    {
      "index": 0,
      "ok": true,
      "value": 0.1,
      "variable": "s1"
    },
    <one entry per expression, in submission order>
  ],
  "variables": {
    "s1": 0.1,
    <one entry per assigned variable>
  },
  "successful_expression_count": 4,
  "invalid_expression_count": 0,
  "tool_usability": [
    <one availability entry per tool>
  ]
}

Current state:
{
  "cumulative_utility": 0,
  "cumulative_probes": 16,
  "remaining_action_batches": 12,
  "pending_action_batch": {
    "batch_id": "batch_01",
    "batch_number": 1,
    "cases": [
      {
        "context_id": "case_001",
        "raw_values": {
          "<field_id>": <value>,
          ...
        }
      },
      <one entry per case in the batch>
    ]
  },
  "tool_usability": [
    <one availability entry per tool, as in the initial state card>
  ]
}
\end{template*}

\end{document}